\documentclass{article}
\usepackage{arxiv}
\usepackage{amsmath}
\usepackage[utf8]{inputenc} 
\usepackage[T1]{fontenc}    
\usepackage{hyperref}       
\usepackage{booktabs}       
\usepackage{amsfonts}       
\usepackage{nicefrac}       
\usepackage{microtype}      
\usepackage{graphicx}
\usepackage{multirow}
\usepackage{longtable}
\usepackage{float}
\usepackage[numbers,sort&compress]{natbib}

\graphicspath{{./}{figs/}}

\title{Interpreting GFlowNets for Drug Discovery: What Probes Can and Cannot Show}

\author{
  Amirtha Varshini A S \\
  Montai Therapeutics \\
  \texttt{asindhanai@montai.com}
  \And
  Duminda S. Ranasinghe \\
  Montai Therapeutics \\
  \texttt{dranasinghe@montai.com}
  \And
  Hok Hei Tam \\
  Montai Therapeutics \\
  \texttt{htam@montai.com}
}

\begin{document}

\maketitle

\begin{abstract}
Generative Flow Networks (GFlowNets) offer a powerful framework for molecular design, yet their internal decision policies remain opaque. This limits adoption in drug discovery, where chemists require interpretable rationales for proposed structures. We introduce an interpretability framework for SynFlowNet~\cite{cretu2025synflownet}, a GFlowNet trained on documented chemical reactions and purchasable starting materials, which confines generation to synthetically accessible chemical space and enables sampling of both viable target molecules and the synthetic routes that produce them. Our approach integrates (i) gradient-based saliency with counterfactual perturbations of molecular substructures, (ii) concept attribution via sparse autoencoders (SAEs) trained on internal embeddings, and (iii) motif probes that assess whether functional groups are decodable from those embeddings. Applied to SynFlowNet, physicochemical properties and functional-group information are readily decodable from its embeddings; however, an architecture-matched untrained control achieves nearly identical physicochemical probe performance, showing that this decodability primarily reflects graph architecture and atom featurization rather than policy training. Illustrative counterfactual edits to high-attribution regions produce small changes in computed QED, but the analyzed checkpoint is unavailable and a saliency-versus-random control remains future work.
We further extend this analysis with an overcomplete BatchTopK sparse autoencoder~\cite{bussmann2024batchtopk}, which exposes fine-grained chemical feature detectors that the compressive (undercomplete) model does not, and we validate the eleven linear probes against descriptor, random-label, scaffold-split, and untrained-network controls, test features by single-feature ablation, and check them for cross-seed stability. The overcomplete model reconstructs the embedding better than a matched undercomplete baseline on identical splits and seeds (NMSE 0.0015 versus 0.0026), with a batch-mean of 64 active features and negligible feature death; single-feature ablation has property-specific effects on probe decoding (specificity ratios of roughly 8 to 23 times the off-target effect for the substantial interventions); and chemically enriched substructure detectors, including separate per-halogen features, emerge in individual runs, while a small subset of dictionary directions is stable across seeds. Our most consequential result is a negative control: a randomly initialized network of identical architecture decodes every probed property essentially as well as the trained policy (drug-likeness within 0.005), so linear decodability of physicochemical properties reflects the graph architecture and atom featurization rather than anything the model learned. We therefore interpret only what survives these controls, not high probe scores as evidence of learned chemical knowledge.
\end{abstract}

\begin{figure}[htbp]
    \centering
    \includegraphics[width=\linewidth,keepaspectratio]{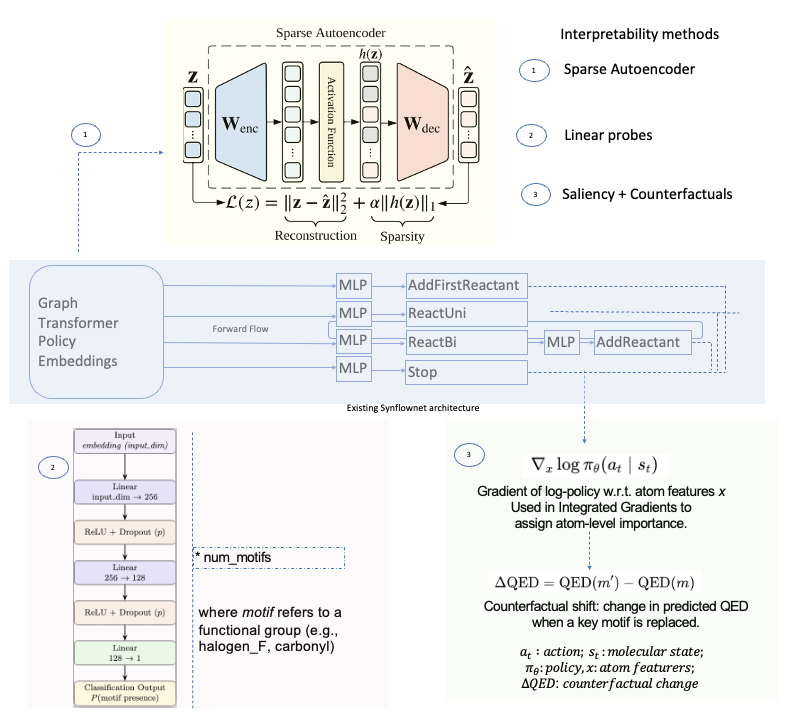}
    \caption{Overview of the proposed interpretability framework for GFlowNets. Our pipeline integrates (1) sparse autoencoders \cite{shu2025surveysparseautoencodersinterpreting} (SAEs) for discovering sparse, property-associated factors such as polarity and lipophilicity, (2) motif probes to test whether embeddings encode functional groups, and (3) gradient-based saliency and counterfactual perturbations for atom- and motif-level attribution. Together, these approaches span fine-grained atomic rationales to high-level medicinal chemistry concepts.}
    \label{fig:methods}
\end{figure}

\section{Introduction}

Drug discovery demands exploration of vast chemical spaces while adhering to strict constraints on synthesizability, drug-likeness, and safety. Although modern deep generative models—including variational autoencoders (VAEs), generative adversarial networks (GANs), and diffusion models—have shown strong capability in proposing novel molecules, they often produce structures that are synthetically infeasible, insufficiently diverse, or justified through opaque decision-making processes. Medicinal chemists, however, require explanations grounded in molecular structure: which functional groups guide the model’s choices, which physicochemical properties are being prioritized, and how small structural modifications affect predicted reward or synthetic feasibility. Without such explanations, even high-performing models remain difficult to trust and integrate into design–make–test–analyze (DMTA) cycles.

Generative Flow Networks (GFlowNets) \cite{bengio2021gflownet, madan2023gflownet} provide a compelling alternative to conventional likelihood- or score-based generative models. Rather than producing molecules in a single denoising or decoding step, GFlowNets learn stochastic policies that construct molecules sequentially, allocating probability mass across a diverse set of high-reward structures. Recent extensions such as SynFlowNet \cite{cretu2025synflownet} incorporate synthesis constraints and reaction-template–based assembly, enabling chemically meaningful and synthetically aware molecule generation. Despite this promise, the interpretability of GFlowNets remains largely unexplored. In contrast to diffusion models—where attribution methods and latent-space analyses have matured—GFlowNets pose unique challenges: the policy must choose among heterogeneous action types, and intermediate graph states contain rich structural information that is rarely interrogated.

Interpreting these internal representations is essential for practical deployment. Medicinal chemists routinely examine how polarity, lipophilicity, aromaticity, and functional-group context influence model predictions, and they seek mechanistic explanations for why a model favors one synthetic route or scaffold over another. Yet current GFlowNet architectures provide little visibility into their decision-making processes, and existing attribution methods are not designed for discrete, graph-based generative policies.

To address this gap, we introduce an interpretability toolkit that adapts methods from supervised learning \cite{sundararajan2017integrated, bau2017network, ying2019gnnexplainer} to structured, sequential, graph-based generative settings. Our approach integrates gradient-based saliency, counterfactual perturbations, sparse autoencoders, and motif-level probes to reveal how SynFlowNet encodes and manipulates chemical information during generation. In doing so, our framework links machine-learning explanations with medicinal-chemistry reasoning and aligns with the growing emphasis on explainability in molecular design \cite{jimenezluna2020explainable, wellawatte2022cf}.

We analyze the representation at two complementary scales. A compressive, \emph{undercomplete} autoencoder ($256\to128$; the \textbf{MLP-SAE}) recovers the dominant physicochemical axes of the embedding, a form of sparse dimensionality reduction close to sparse PCA~\cite{zou2006sparse}. An \emph{overcomplete} autoencoder ($256\to1024$) instead recovers fine-grained feature detectors, including those held in superposition~\cite{elhage2022superposition}, in the sense established by Bricken et al.~\cite{bricken2023monosemanticity} and Cunningham et al.~\cite{cunningham2023sparse} for language models. Throughout, we treat validation as a first-class concern: for every linear probe we run descriptor and random-label control tasks~\cite{hewitt2019control}, we ablate individual features to test for specificity, and we retrain across seeds to test which features reproduce. This separates genuinely learned structure from information that is trivially available in the molecular graph, and it lets us report what the analysis does and does not license.

\paragraph{Data provenance.} All analyses use frozen graph-level embeddings from a single SynFlowNet checkpoint trained with QED reward. That checkpoint no longer exists; the extracted embeddings (32{,}054 molecules) are the surviving, reproducible artifact, and retraining would not reproduce the same weights. We therefore deposit the embeddings alongside the code (Appendix~\ref{sec:availability}), and all results below are computed from them.

\section{Methods}
In our experiments, we analyze SynFlowNet trained with QED (Quantitative Estimate of Drug Likeness)~\cite{bickerton2012qed} as the reward function. Our interpretability framework combines three complementary approaches, each targeting a different level of explanation. Gradient based saliency identifies the atoms and bonds that influence specific generative actions, providing fine grained local attribution. Counterfactual perturbations extend this by testing the sensitivity of action probabilities and reward outcomes to structured molecular edits \cite{wellawatte2022cf, lucic2022cfgexplainer, karimi2020algorithmicrecoursecounterfactualexplanations}. Sparse autoencoders (SAEs) recover sparse, axis aligned latent factors, enabling analysis of how abstract representations relate to physicochemical properties \cite{bricken2023monosemanticity, cunningham2023sparse}. 
Finally, motif probes test whether discrete chemical motifs are encoded in the learned embeddings, linking internal representations to recognizable medicinal chemistry concepts. Together, these methods provide multi scale interpretability, spanning atom level rationales, latent space structure, and motif level detectors.

\subsection{Gradient-Based Saliency with Counterfactual QED Analysis}

We estimate atom-level saliency for SynFlowNet by applying integrated gradients (IG) \cite{sundararajan2017integrated} to the log-probability of the \texttt{Stop} action.
Given an input molecular state $s_t$ represented by atom features $x$ and a baseline $\tilde{x}$ (either a zero vector or the mean feature vector), we approximate the IG attribution for atom feature $x_i$ on the \texttt{Stop} log-probability as \[
\mathrm{IG}_i(\text{Stop}, s_t) \approx (x_i - \tilde{x}_i)
\frac{1}{M} \sum_{m=1}^M
\nabla_{x_i} \log \pi_\theta(\text{Stop} \mid s_t^{(m)}),
\]
where $s_t^{(m)}$ linearly interpolates between $\tilde{x}$ and $x$ and $M=64$ steps are used in practice. In our implementation, we obtain logits from SynFlowNet, apply a \texttt{log\_softmax} over the action dimension, and backpropagate gradients from the \texttt{Stop} log-probability to the atom features. Atom-level importance scores are then computed by aggregating the absolute attributions across feature dimensions.

We focus on the \texttt{Stop} action because it is the point in generation where the entire molecule is present, and the model decides that the structure is complete. This allows us to attribute importance to the full molecular graph. A current limitation of this choice is that it provides saliency only at the final decision step and does not capture how intermediate states influence earlier action probabilities. Extending this analysis to other decision points in the trajectory is an important direction for future work.

To move beyond purely gradient-based attributions, we construct motif-level explanations and evaluate their effect on drug-likeness. Given a molecule, we first threshold atom scores at the 75th percentile, extract connected components of high-saliency atoms, and augment these with ring systems detected by RDKit (via \texttt{GetSymmSSSR}) as candidate
motifs. Each candidate motif is scored by the sum of its atom attributions (with a slight boost for rings), and we select the top-$k$ non-overlapping motifs.

We then perform counterfactual edits targeted to these motifs using a fixed set of chemically motivated RDKit transformation rules (e.g., ether~$\rightarrow$~thioether, methyl~$\rightarrow$~fluorine, chloro~$\rightarrow$~bromo, amide~$\rightarrow$~ester).
For each selected motif, we apply all compatible transformations, sanitize the resulting molecules, and compute QED using RDKit. This yields a set of valid counterfactuals with associated reward changes $\Delta \mathrm{QED} = \mathrm{QED}(m') - \mathrm{QED}(m)$.
For each motif, we report the best counterfactual edit (if any) and its corresponding $\Delta \mathrm{QED}$, providing an intervention-based view of how salient motifs contribute to and can be modified to improve drug-likeness. The integration step count ($M=64$) and the six transformation rules are fixed in the released code; the two illustrative $\Delta\mathrm{QED}$ values reported in Section~\ref{sec:results} are from the original run and could not be independently re-verified, because computing new saliency requires a forward pass through the trained policy and the analyzed checkpoint no longer exists (Appendix~\ref{sec:availability}).

\subsection{Undercomplete Sparse Autoencoder (MLP-SAE)}
To analyze whether SynFlowNet embeddings capture chemically meaningful structure, we first apply a compressive, undercomplete sparse autoencoder, which we refer to as the \textbf{MLP-SAE}.
Its code is smaller than the input. Given a hidden embedding $h \in \mathbb{R}^d$ with $d=256$ from the policy network, it encodes a nonnegative sparse code $z \in \mathbb{R}^{128}$ and reconstructs $\hat h \in \mathbb{R}^{256}$ through a four-layer nonlinear MLP autoencoder with ReLU and dropout and a $128$-dimensional bottleneck:
\begin{align*}
z      &= \mathrm{ReLU}\!\big(W_2\,\mathrm{Dropout}(\mathrm{ReLU}(W_1 h + b_1)) + b_2\big),\\
\hat h &= W_4\,\mathrm{Dropout}(\mathrm{ReLU}(W_3 z + b_3)) + b_4,
\end{align*}
with $W_1\in\mathbb{R}^{256\times256}$, $W_2\in\mathbb{R}^{128\times256}$, $W_3\in\mathbb{R}^{256\times128}$, $W_4\in\mathbb{R}^{256\times256}$. Training minimizes
\[
\mathcal{L}=\|h-\hat h\|_2^2+\lambda \|z\|_1,
\]
where the $\ell_1$ term on the bottleneck code encourages sparse, axis aligned factors. 
In practice, we correlate each factor with molecular descriptors (e.g., TPSA for polarity, Crippen logP for lipophilicity) to test whether these abstract dimensions map to interpretable physicochemical axes.

We train the MLP-SAE on frozen SynFlowNet embeddings extracted from the final graph transformer layer after pooling over atoms for each molecular state. The bottleneck (code) dimension is 128 (determined after experimentation) and the dropout rate is 0.1. The model is optimized with Adam (learning rate $1\times10^{-3}$) for 200 epochs and a batch size of 128. We impose sparsity using an $\ell_1$ penalty with coefficient $\lambda = 0.01$ and target sparsity 0.05. Training is performed on the full set of molecular embeddings, with held out data reserved for evaluation as detailed in Appendix~\ref{sec:sae_analysis}; the train/held-out split uses \texttt{random\_state}$=42$ and the SynFlowNet policy was trained with seed $0$.
For downstream analysis, we additionally train a three-layer MLP reward predictor for 100 epochs using the same learning rate and a dropout rate of 0.2. Because this model is undercomplete, it characterizes global structure but does not support feature-level (monosemanticity) claims; those require the overcomplete dictionary of Section~\ref{sec:overcomplete}.

\subsection{Motif Probes}
Whereas SAEs identify continuous latent factors, motif probes test whether discrete chemical motifs are encoded in SynFlowNet embeddings. We freeze the pretrained GFlowNet and train shallow feedforward classifiers on the embeddings $h$ to predict motif presence. Labels are obtained automatically using RDKit SMARTS pattern matching for functional groups, aromatic rings, and halogens. High probe performance indicates that motif information is accessible in the embeddings, linking abstract representations back to recognizable medicinal chemistry concepts.

For motif probing, we freeze SynFlowNet and extract the same pooled embeddings $h$ as in the SAE analysis. For each SMARTS-defined motif $m$, we train a separate feedforward classifier to predict motif presence from $h$. The probe is a three-layer MLP with architecture $256 \rightarrow 256 \rightarrow 128 \rightarrow 64 \rightarrow 1$, using ReLU activations and a  dropout rate of 0.2 after each hidden layer. Before training, embeddings are standardized with a 
\texttt{StandardScaler}. Each probe is optimized using Adam (learning rate $1\times 10^{-3}$) with a binary cross-entropy loss (via \texttt{BCEWithLogitsLoss}) for 50 epochs under class balanced
sampling. We evaluate performance on a held-out test split using AUROC and average precision (AP), as detailed in Appendix~\ref{sec:motif_appendix}.

\subsection{Overcomplete BatchTopK Sparse Autoencoder}
\label{sec:overcomplete}
The MLP-SAE identifies which major axes the model uses. To ask instead which fine-grained features the model represents, including features distributed in superposition, we train an \emph{overcomplete} sparse autoencoder~\cite{olshausen1997sparse} with a $4\times$ expansion. Given $h \in \mathbb{R}^{256}$ it encodes $z \in \mathbb{R}^{1024}$ and reconstructs $\hat h \in \mathbb{R}^{256}$:
\[
z_{\text{pre}} = \mathrm{ReLU}(W_{\text{enc}}(h - b_{\text{pre}}) + b_{\text{enc}}), \qquad
z = \mathrm{BatchTopK}(z_{\text{pre}}, k), \qquad
\hat h = W_{\text{dec}} z + b_{\text{dec}},
\]
with $W_{\text{enc}}\in\mathbb{R}^{1024\times256}$, $W_{\text{dec}}\in\mathbb{R}^{256\times1024}$, $b_{\text{pre}}$ initialized to the training mean, and $k=64$. BatchTopK~\cite{bussmann2024batchtopk} keeps exactly $k\times B$ activations across a batch of $B$ samples, so the per-batch mean sparsity is exactly $k$ while the per-molecule count varies; no $\ell_1$ coefficient is tuned. Decoder columns are renormalized to unit norm after each step, and the objective is mean squared reconstruction error only. Dead features (features not firing in the preceding epoch) are resampled every 10 epochs from epoch 20, capped at 2\% per round. We train with Adam (learning rate $3\times10^{-4}$), batch size 512, 50 epochs, a 90/10 split, standardized inputs, and five seeds (42 to 46). Unit tests assert that the batch-mean $L_0=k$ and that decoder columns are unit norm.

\subsection{Linear Probes and Trivial Baselines}
\label{sec:probes}
To measure information content we use strictly linear probes (Ridge regression and logistic regression), in contrast to the MLP probes above. We probe the raw embedding $h$, the sparse latent $z$, and the reconstruction $\hat h$, for eleven targets (Appendix~\ref{sec:targets}). We add two control tasks~\cite{hewitt2019control}. The first shuffles labels (\texttt{random\_label}); a sound probe collapses to chance. The second runs the identical probe on RDKit counts plus ECFP4 fingerprints with no model involved (\texttt{descriptor}). The descriptor baseline comprises count features and fingerprints and does not include MolWt, LogP, or TPSA directly. Where a target equals one of the count columns, we drop that column per target so the comparison is informative rather than arithmetic: flexibility (defined as NumRotatableBonds$/10$), the complexity ring-proxy (its ring-count constituents), and the halogen, boron, and aromaticity classification targets (their atom- and ring-count columns). For molecular weight, size, polarity, lipophilicity, drug-likeness, and urea no column is dropped, because their defining descriptors are correlated with but not present among the baseline features; we report the leakage-free baseline throughout. Probes use an 80/10/10 split and probe seeds 0 to 4. The released code reproduces the reported configuration, including the logistic-regression iteration limit behind Figures~\ref{fig:f1}--\ref{fig:f10} and \texttt{NUMBERS\_MANIFEST.csv}. Under that limit the logistic-regression probes do not always reach the default convergence tolerance; all classification AUROCs are $\geq 0.92$.

\subsection{Feature Characterization, Ablation, and Stability}
For each target we rank overcomplete features by $|w|$ from the linear probe. To characterize a feature we compare its 30 highest-activating molecules against 30 silent molecules, using RDKit descriptors (Mann-Whitney U) and SMARTS prevalence (Fisher exact), with Benjamini-Hochberg correction; a feature is enriched when the prevalence gap exceeds 30 points at $q<0.01$. For an interventional check at the level of probe decoding we encode, zero one feature, decode, and re-apply the frozen probes, then measure the change per target. This is ablation only: nothing is amplified and generation is never re-run. For stability we retrain across seeds and match decoder columns by cosine similarity against a chance baseline from random dictionaries.

\section{Results}
\label{sec:results}

\begin{figure}[htbp]
    \centering
    \includegraphics[width=\linewidth,keepaspectratio]{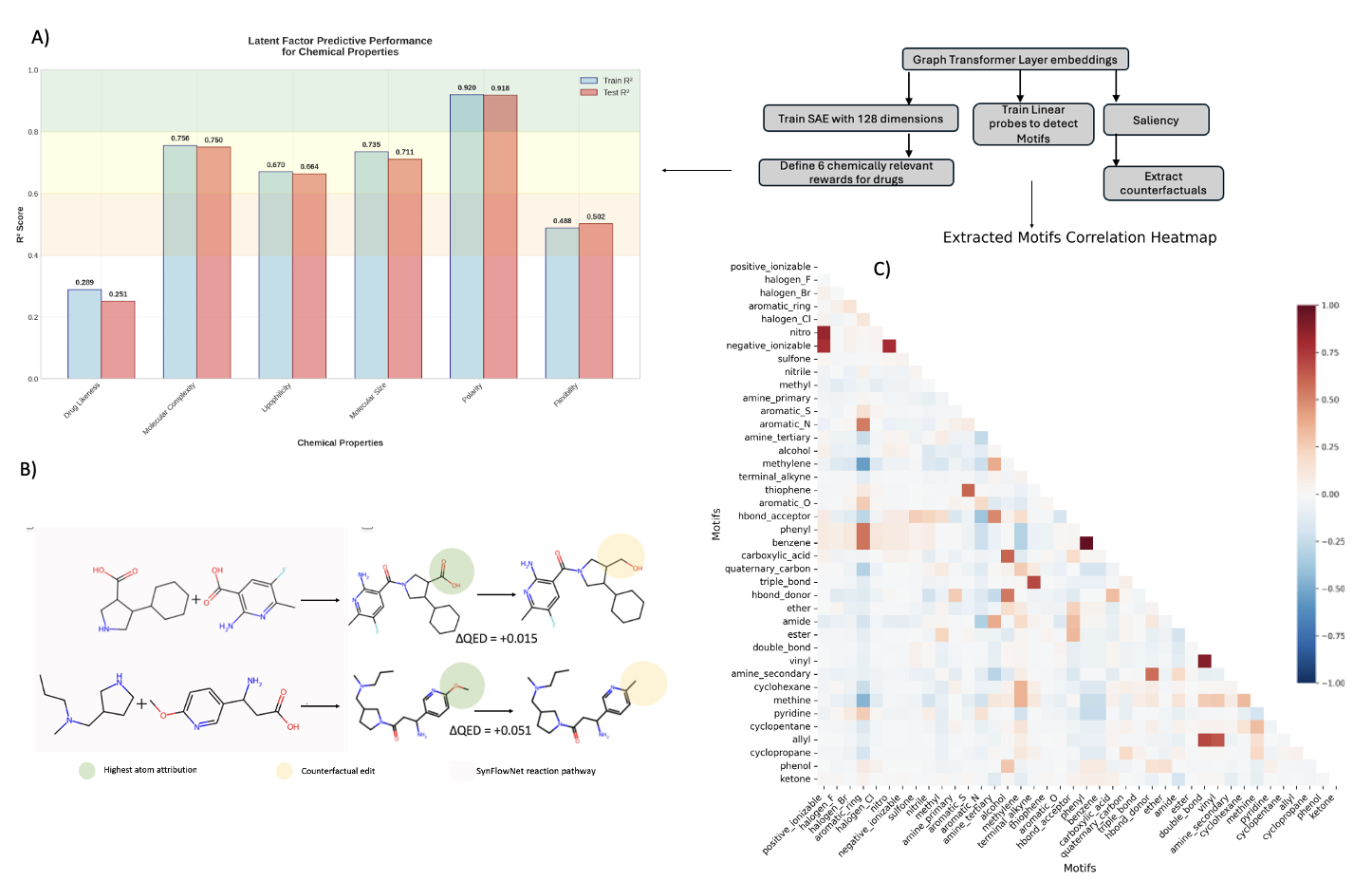}
    \caption{Interpretability results on SynFlowNet embeddings. (A) Predictive performance of an MLP trained on undercomplete (MLP-SAE) factors across six chemical properties, showing that factors are more predictive of polarity, size, and lipophilicity than of composite QED. (B) Example SynFlowNet trajectories with their atom-level saliency (highlighted atoms) and a counterfactual edit that alters predicted QED, illustrating intervention-based attribution. (C) Motif–motif correlation heatmap from motif probes (both axes are motifs), showing that co-occurring functional groups such as halogens and aromatic rings cluster.}
    \label{fig:results}
\end{figure}

\paragraph{Counterfactual saliency.}
Integrated Gradients applied to the \texttt{Stop} log probability produces atom level saliency maps that consistently highlight chemically meaningful regions of the molecule. Aggregating high scoring atoms into connected components and ring systems yields motif candidates that align with polar substituents, aromatic fragments, and other regions that influence reward (Fig.~\ref{fig:results}B). Targeted RDKit based counterfactual edits within these motifs produce shifts in QED (for example, $\Delta\mathrm{QED}=+0.015$ and $+0.051$ for the examples in Fig.~\ref{fig:results}B), with transformations such as halogen exchange, ester or amide modification, and oxidation state changes. We observe a tendency for higher-saliency motifs to yield larger reward changes; these effect sizes are small relative to QED's range, and a controlled comparison against edits to randomly chosen motifs is needed to establish the relationship quantitatively, which we leave for future work.

\paragraph{QED related latent structure.}
Training the MLP-SAE on SynFlowNet embeddings yields 128 latent factors with a mean activation sparsity of 0.105, corresponding to the fraction of molecules for which each factor is active. A factor is considered active when its nonnegative activation value $z_i$ is greater than a small threshold, indicating that the SAE has identified a meaningful direction in the embedding space for that molecule. Several factors show strong linear associations with physicochemical properties. For example, Factor 11 correlates with molecular size ($r = 0.76$), while Factor 86 and Factor 118 capture complementary aspects of polarity ($r = -0.57$ and $r = 0.54$). An MLP predictor trained on the SAE factors achieves high $R^2$ values for polarity (0.92) and size (0.71), much higher than for the composite QED score (0.25). These findings indicate that reward-relevant properties are organized in the embedding along more predictable axes such as polarity, lipophilicity, and size, even though QED itself is a nonlinear combination of these components; Section~\ref{sec:randomnet} indicates this organization is largely architecture-driven rather than a product of training.

\paragraph{Motif decoding.}
Motif probes trained on pooled embeddings exhibit strong classification performance across a wide range of functional groups. Halogens, aromatic ring systems, and carbonyl containing motifs achieve AUROC scores above 0.9 (Table~\ref{tab:motif_results} in Appendix~\ref{sec:motif_appendix}), even though the probes are shallow feed forward networks. This shows that SynFlowNet embeddings encode chemically interpretable substructures that shallow classifiers can recover. For the highest-scoring motifs, however, near-perfect AUROC may reflect atom-type information that is directly available in the graph rather than a learned abstraction; the trivial-baseline controls in Section~\ref{sec:decodability} quantify this.

\subsection{Overcomplete reconstruction and sparsity}
\begin{figure}[htbp]
    \centering
    \includegraphics[width=\linewidth,keepaspectratio]{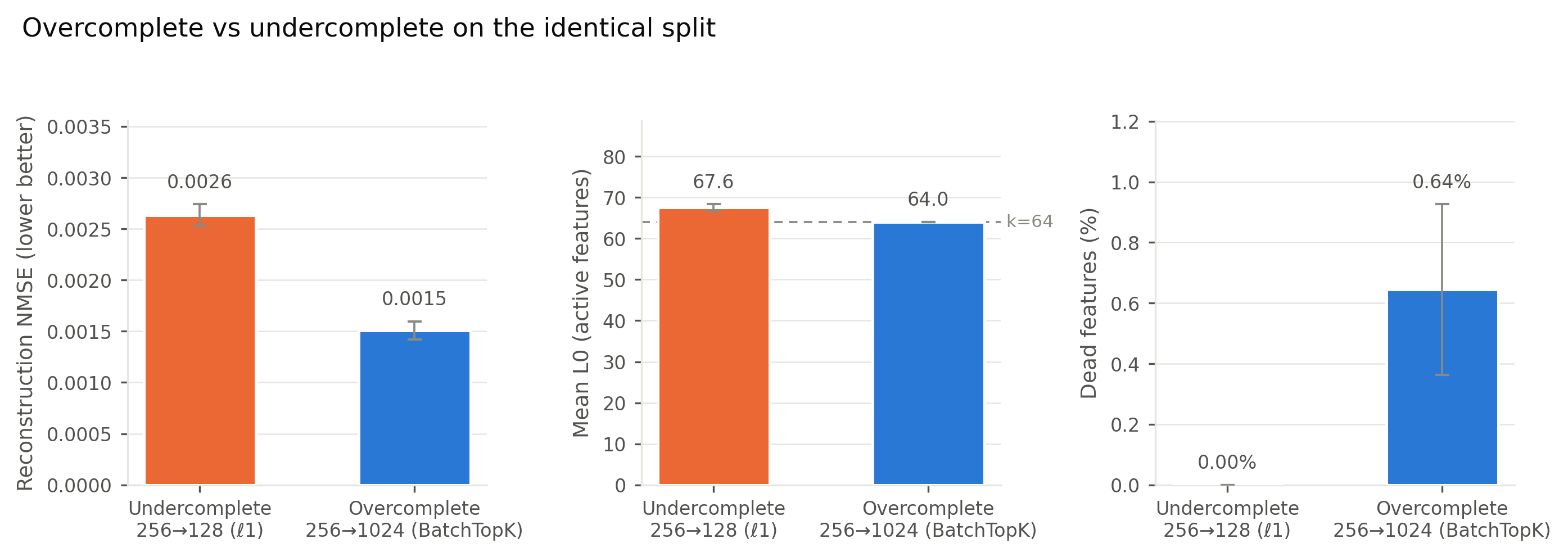}
    \caption{Overcomplete versus a matched linear baseline on the identical split, seeds, and data. Left: reconstruction NMSE (lower is better). Middle: mean active features. Right: dead-feature fraction. The overcomplete model reconstructs 1.75 times better at comparable sparsity.}
    \label{fig:f3}
\end{figure}

\begin{figure}[htbp]
    \centering
    \includegraphics[width=\linewidth,keepaspectratio]{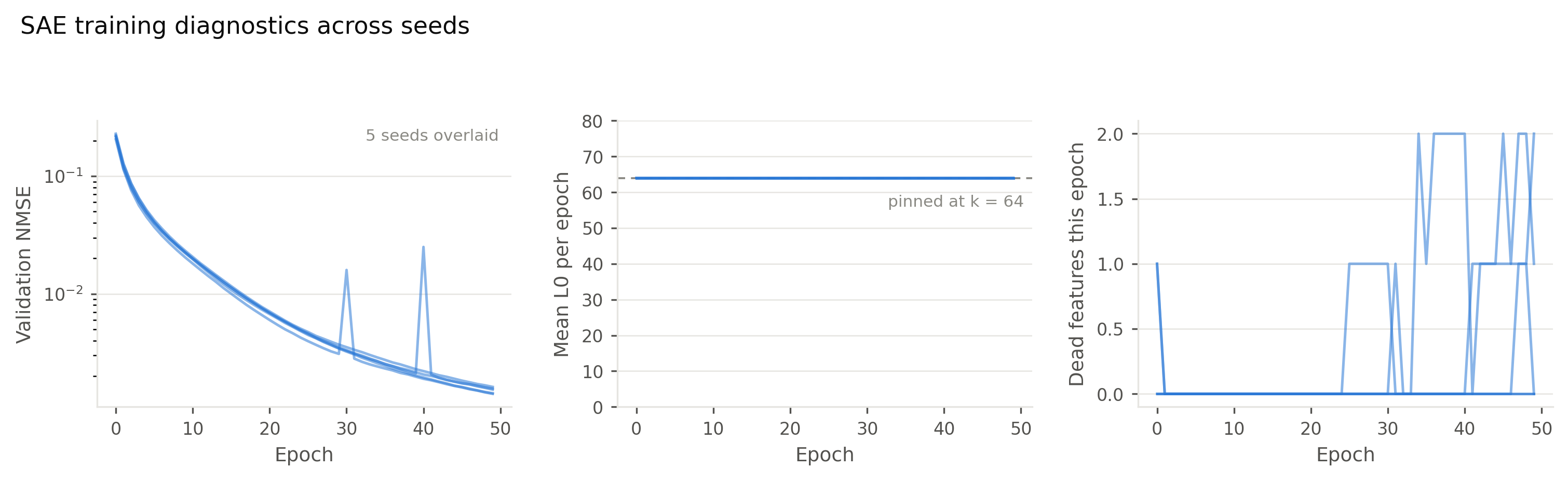}
    \caption{Overcomplete SAE training diagnostics across five seeds. Left: validation NMSE decreases smoothly (brief spikes are dead-feature resampling events that recover within a few epochs). Middle: batch-mean $L_0$ is pinned at $k=64$. Right: dead features stay near zero under periodic resampling.}
    \label{fig:f5}
\end{figure}

The overcomplete BatchTopK autoencoder reconstructs the embedding at NMSE $0.00151 \pm 0.00009$, with a batch-mean of exactly 64 active features per example ($L_0 = 64.000 \pm 0.000$ across seeds; per-molecule counts vary) and negligible feature death ($0.64\% \pm 0.28\%$). On the identical split and seeds, a \emph{matched linear baseline} ($256 \to 128$, single linear layer each way, $\ell_1$) reaches NMSE $0.00263 \pm 0.00011$ (Fig.~\ref{fig:f3}), a $1.75\times$ worse reconstruction. The two models differ in both width and sparsity mechanism, but the matched baseline reaches batch-mean $L_0 = 67.6$ on its own, comparable to $k=64$, so at similar average activity the tested $1024$-dimensional BatchTopK model reconstructs substantially better than the tested $128$-dimensional $\ell_1$ model. Training is stable across seeds, batch-mean $L_0$ is pinned at 64 every epoch, and dead features stay near zero (Fig.~\ref{fig:f5}). This matched linear baseline is distinct from the four-layer MLP-SAE used for the factor analysis above.

\subsection{Information preservation through the bottleneck}
\begin{figure}[htbp]
    \centering
    \includegraphics[width=0.85\linewidth,keepaspectratio]{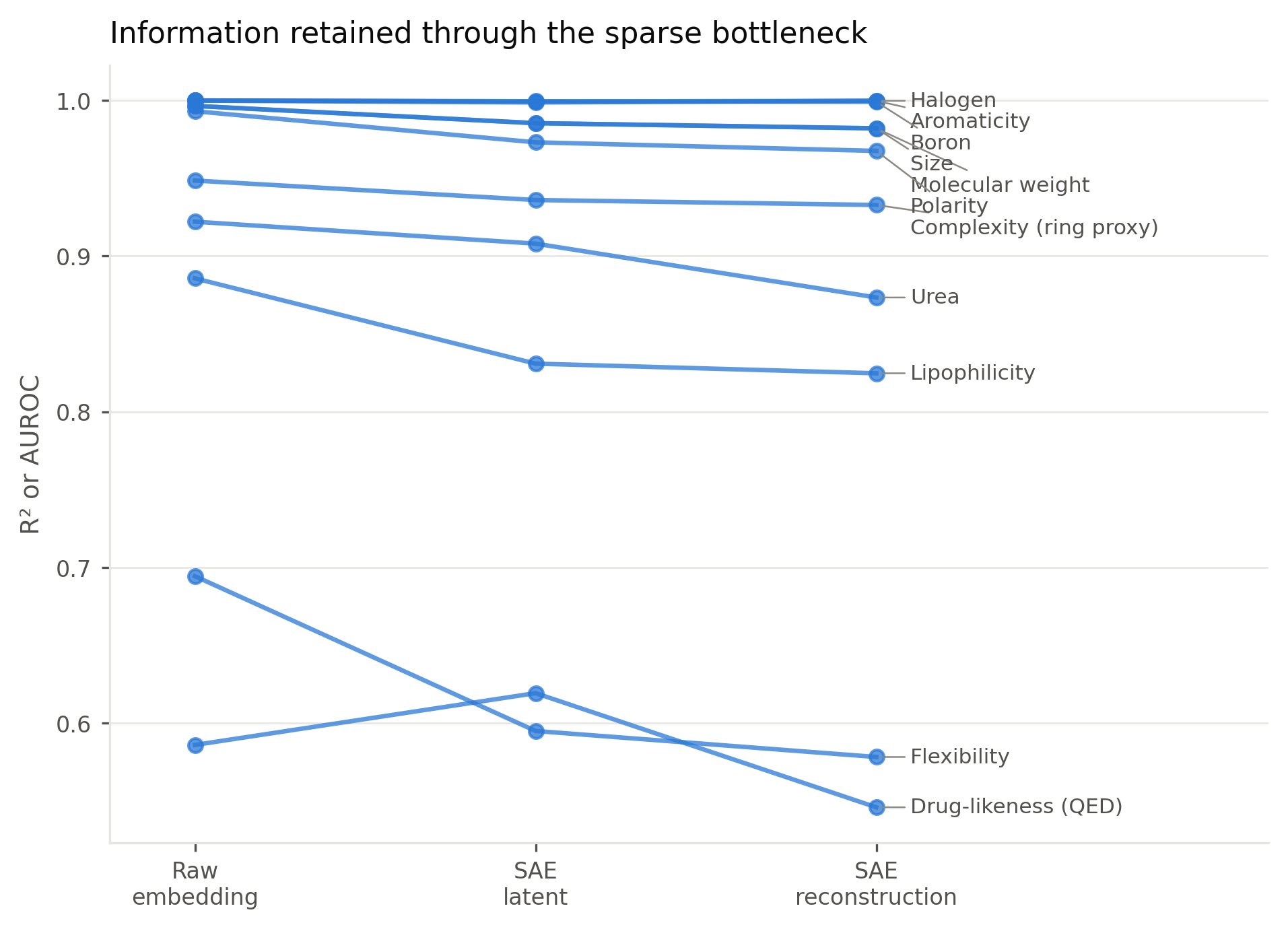}
    \caption{Information retained through the sparse bottleneck: linear-probe performance on the raw embedding, the sparse latent, and the reconstruction. Most targets change little; drug-likeness (QED) rises from raw to latent.}
    \label{fig:f4}
\end{figure}

Passing test embeddings through the encode-decode pipeline and re-applying the pre-trained linear probes shows a median retention of 98.3\% (Fig.~\ref{fig:f4}). Classification is essentially untouched (change $\leq 0.001$ for halogen, aromaticity, and boron). Retention is not uniform, however: the worst case is flexibility, which loses $\Delta R^2 = -0.116$ from a base of 0.694, a 17\% relative drop (83\% retained). One target behaves in the opposite direction: drug-likeness is decoded better from the sparse latent than from the raw embedding ($\Delta R^2 = +0.033$), so the sparse basis makes QED more linearly accessible than it is in the dense embedding.

\subsection{Linear decodability and trivial baselines}
\label{sec:decodability}
\begin{figure}[htbp]
    \centering
    \includegraphics[width=0.9\linewidth,keepaspectratio]{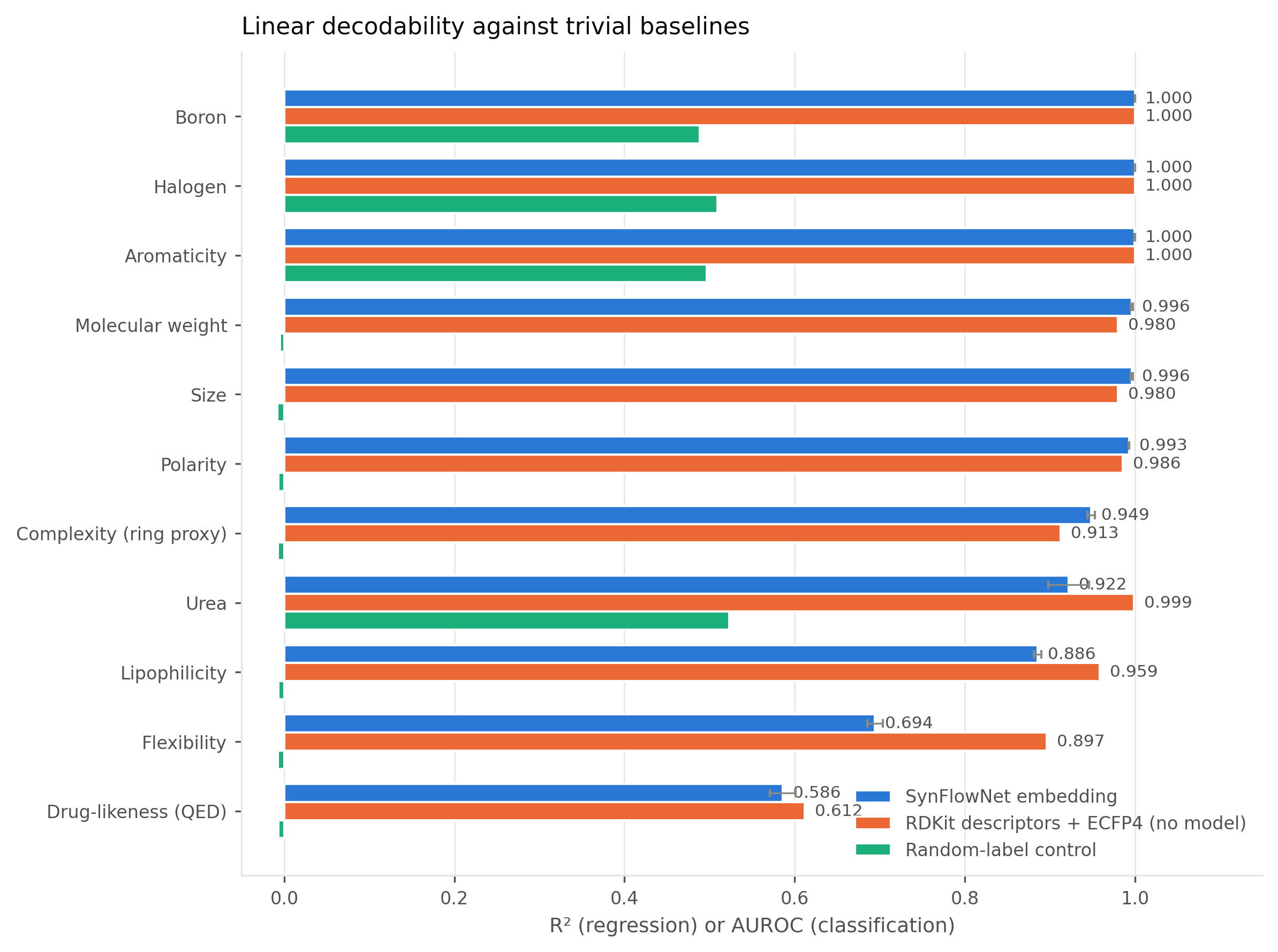}
    \caption{Linear decodability against trivial baselines, per target. Blue: SynFlowNet embedding. Orange: RDKit descriptors plus ECFP4, no model. Green: random-label control. Random-label sits at chance everywhere; under this random split the embedding never exceeds the leakage-free descriptor baseline by more than $+0.036$ and is worse on four targets.}
    \label{fig:f1}
\end{figure}

\begin{figure}[htbp]
    \centering
    \includegraphics[width=0.85\linewidth,keepaspectratio]{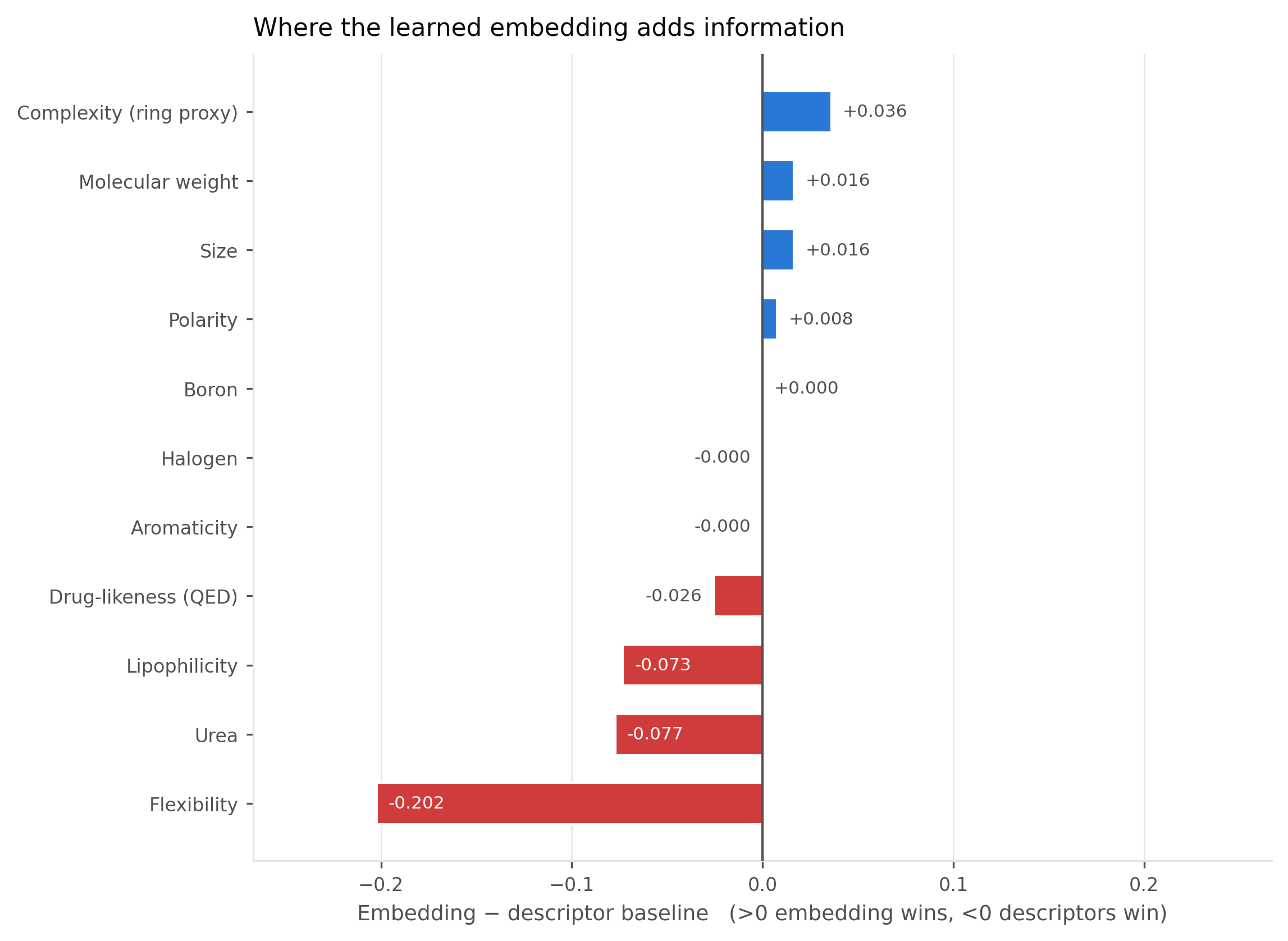}
    \caption{Signed gap between the embedding and the leakage-free descriptor baseline under the random split. Right of zero, the embedding adds information; left of zero, descriptors win. No gap exceeds $+0.036$, and four are negative (largest: flexibility, $-0.202$).}
    \label{fig:f2}
\end{figure}

In absolute terms the embedding decodes chemistry well: QED at $R^2=0.59$, size at $0.996$, halogen at AUROC $1.000$, with random-label controls at chance (AUROC 0.49 to 0.52), so the scores are real and not memorization (Fig.~\ref{fig:f1}). We first report the comparison against descriptors under a random split (Fig.~\ref{fig:f2}). On no target does the embedding exceed the leakage-free RDKit descriptor baseline by more than $+0.036$ (the complexity proxy); the next-largest advantages are molecular weight and size ($+0.016$ each) and polarity ($+0.008$); and on four targets the descriptor baseline is higher, by $0.026$ (QED), $0.073$ (lipophilicity), $0.077$ (urea), and $0.202$ (flexibility). Under a random split the learned embedding therefore provides no substantial advantage over cheap descriptors. This ordering, however, depends on the split (Section~\ref{sec:scaffold}), and neither ordering should be read as evidence of learned chemistry, because an untrained network of the same architecture reproduces it (Section~\ref{sec:randomnet}). Every probed property is a deterministic function of the molecular graph that RDKit computes directly, so a descriptor baseline was always going to be strong. We report these comparisons to prevent overclaiming; what the embedding is good at is its own task, sequential action selection, which no probe here tests.

Three points on interpretation. First, size and molecular weight are effectively the same variable here ($\text{size}=\mathrm{clamp}(\mathrm{MolWt}/500,0,2)$, a clip that almost never binds), and both score identically; we retain both only for comparability with the earlier analysis and do not count them as independent evidence. Second, the linear $R^2=0.59$ for QED on the raw embedding exceeds the $0.25$ obtained by a nonlinear MLP on the undercomplete factors (above); the gap reflects information discarded by the $256\to128$ compressive bottleneck, not a failure of linear probing. The descriptor baseline for QED is itself only $0.61$, well below 1.0 despite QED being a deterministic function of RDKit descriptors, because a linear probe cannot capture QED's nonlinear desirability functions. Third, the one target where the embedding leads, the complexity proxy, is the weakest place to hang a positive result: it is a clipped two-ring-count expression whose value \emph{decreases} with ring count and whose effective range is narrow, so we do not lean on it, and we flag a genuine synthetic-accessibility or Bertz-complexity target as a better test for future work.

\subsection{Robustness to scaffold splitting}
\label{sec:scaffold}
\begin{figure}[htbp]
    \centering
    \includegraphics[width=\linewidth,keepaspectratio]{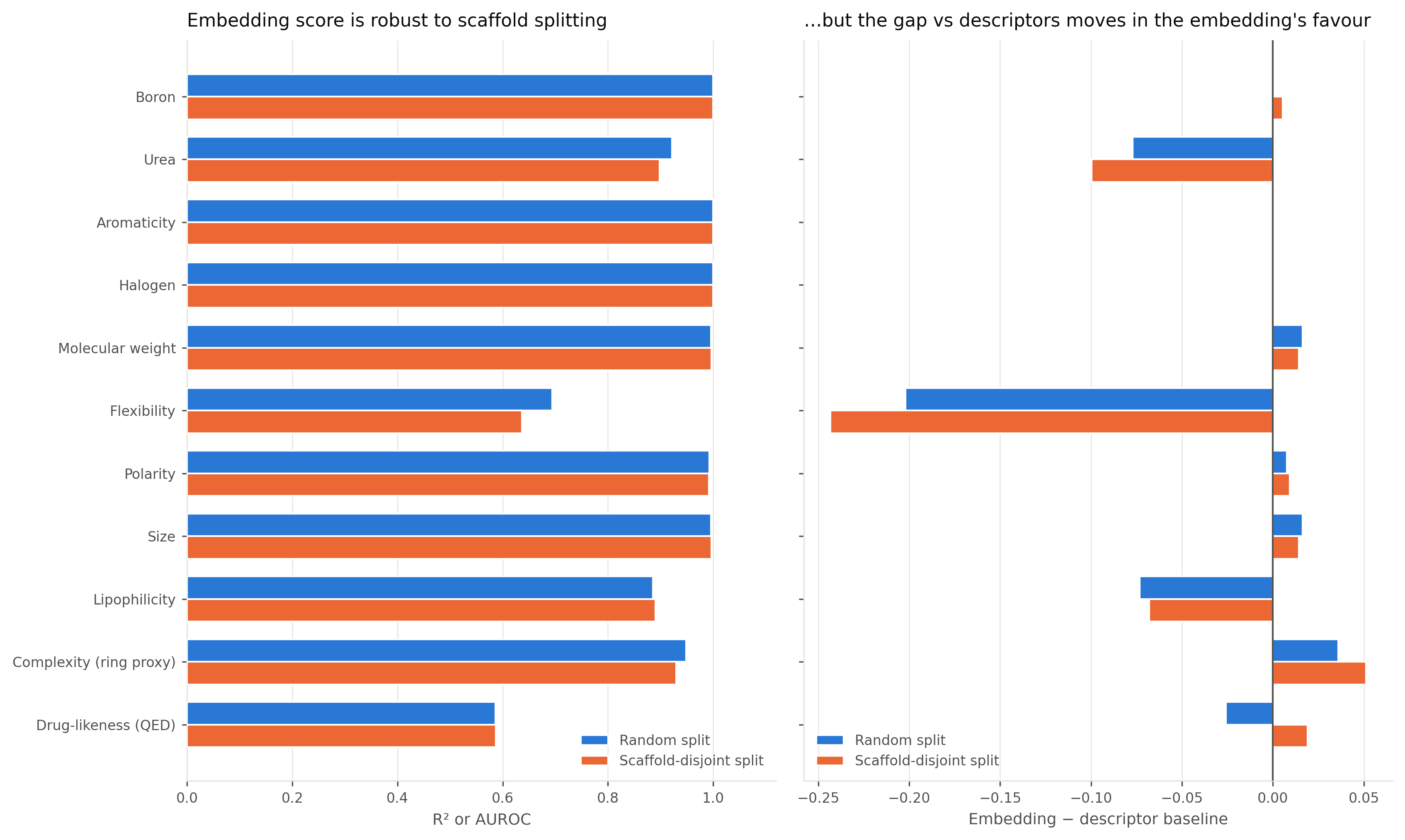}
    \caption{Random versus scaffold-disjoint split. Left: embedding scores are essentially unchanged by the harder split. Right: the signed gap versus descriptors moves in the embedding's favour, because fingerprints rely on scaffold memorization more than the embedding does.}
    \label{fig:f9}
\end{figure}

Molecules sampled from a single generative policy are heavily scaffold-correlated, so a random train/test split risks leaking near-duplicate scaffolds across the boundary and inflating every score. We therefore repeated every probe under a Bemis-Murcko scaffold-disjoint split~\cite{bemis1996scaffold} (18{,}650 distinct scaffolds; groups assigned largest-first to train following the DeepChem convention, so the 3{,}206 test molecules are all scaffold singletons and hold the rarest chemistry, with zero train-test scaffold overlap; tie-broken group order is shuffled per seed, giving five genuinely different scaffold-disjoint splits). The embedding scores barely move (Fig.~\ref{fig:f9}, left): the largest change is flexibility ($-0.057$), and drug-likeness moves by $+0.001$. So scaffold leakage was not inflating the embedding numbers. The descriptor baseline, however, does move: the QED descriptor score falls from 0.612 to 0.568, which flips the gap from $-0.026$ to $+0.019$ (Fig.~\ref{fig:f9}, right). Under the harder split the embedding loses on only three of eleven targets (flexibility, urea, lipophilicity); it is ahead on six and exactly tied on two (halogen and aromaticity, both 1.000 versus 1.000). ECFP4 leans on scaffold memorization more than the embedding does, so the random split flatters the descriptors. As the next result argues, however, even this scaffold-split advantage is likely not evidence of learning.

\subsection{Does training matter? An untrained-network control}
\label{sec:randomnet}
The controls above show the probes are sound and not memorizing, but they do not separate what the \emph{trained} policy contributes from what the architecture and atom featurization provide for free. To test this directly, we instantiated the identical architecture (\texttt{GraphTransformerSynGFN}, 4 layers, 2 heads, Morgan-1024 features, 2.9M parameters), left it \textbf{untrained at random initialization}, extracted embeddings for the same 32{,}054 molecules, and ran the identical linear probes. This control needs only the architecture, not a checkpoint, so it was available even though the analyzed weights are gone.

\begin{figure}[htbp]
    \centering
    \includegraphics[width=0.9\linewidth,keepaspectratio]{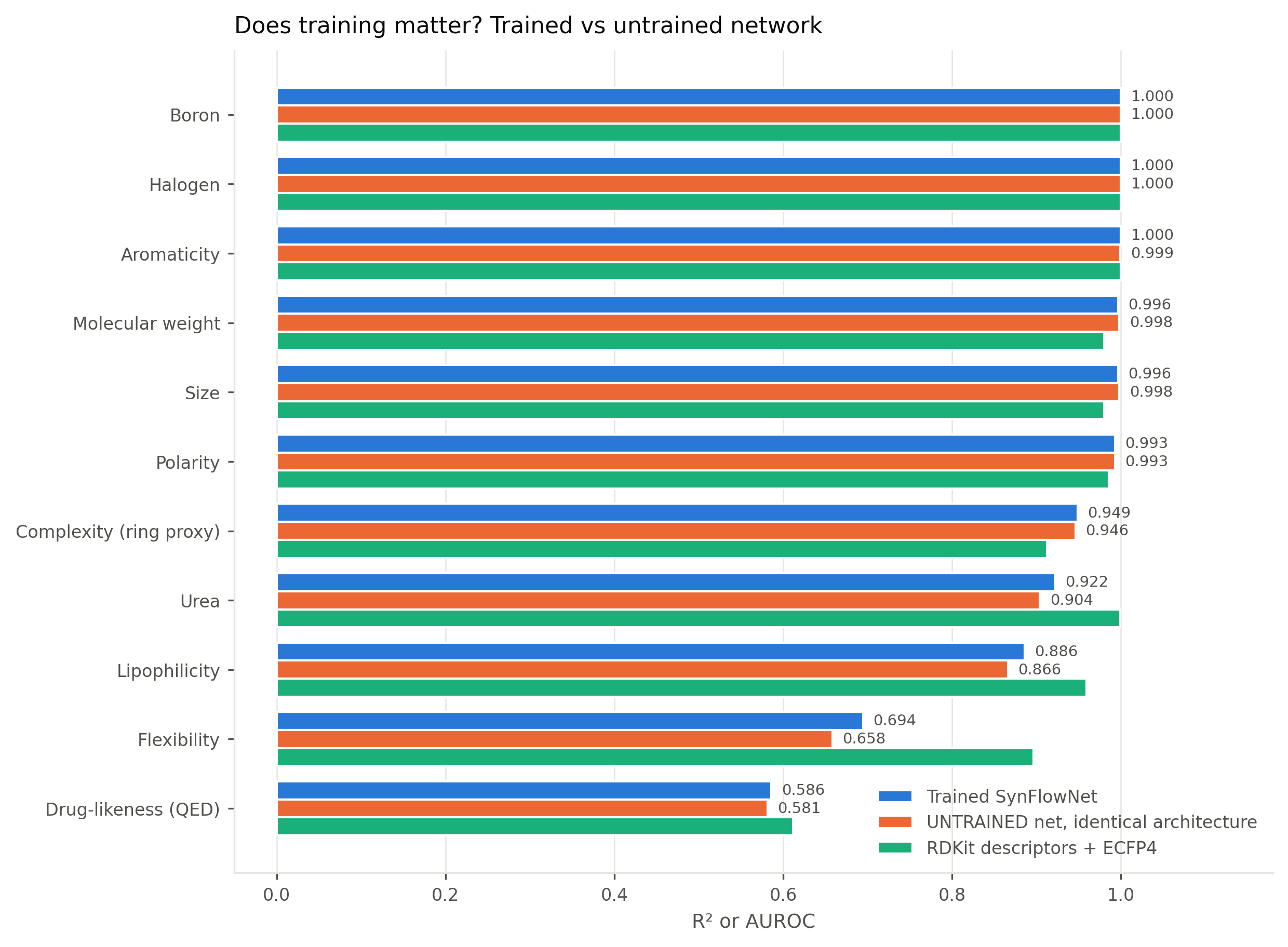}
    \caption{Trained SynFlowNet versus an untrained network of identical architecture, with the RDKit descriptor baseline for reference. The untrained network matches the trained one on every target (largest gain from training: flexibility, $+0.036$; drug-likeness, $+0.005$), so linear decodability reflects architecture and featurization rather than learned weights.}
    \label{fig:f10}
\end{figure}

Training contributes almost nothing to linear decodability (Fig.~\ref{fig:f10}). The untrained network decodes every property to within 0.036 of the trained policy; on the headline property, drug-likeness, training buys $+0.005$ ($0.581 \pm 0.016$ untrained versus $0.586$ trained), well inside the seed spread; and on molecular size and weight the untrained network is marginally \emph{better}. A randomly initialized graph transformer over the same atom featurization already produces embeddings from which these properties are linearly decodable. Linear decodability of physicochemical properties therefore reflects the graph architecture and input representation, not learned structure, and we do not interpret high probe scores, regardless of split, as evidence of acquired chemical knowledge. This also cautions against interpreting the scaffold-split advantage of Section~\ref{sec:scaffold} as evidence of training: the random-split control shows the trained-versus-untrained gap is negligible, though an architecture-matched untrained control under the scaffold split remains to be tested. The comparison there is best read as graph features versus fingerprints, not as learning. Training obviously did something (the trained policy generates high-reward synthesizable molecules and the untrained one cannot); the correct conclusion is that this family of probes is insensitive to what training accomplished, so it should not be used as evidence of learned chemistry.

\subsection{Feature-level interpretation}
\begin{figure}[htbp]
    \centering
    \includegraphics[width=\linewidth,keepaspectratio]{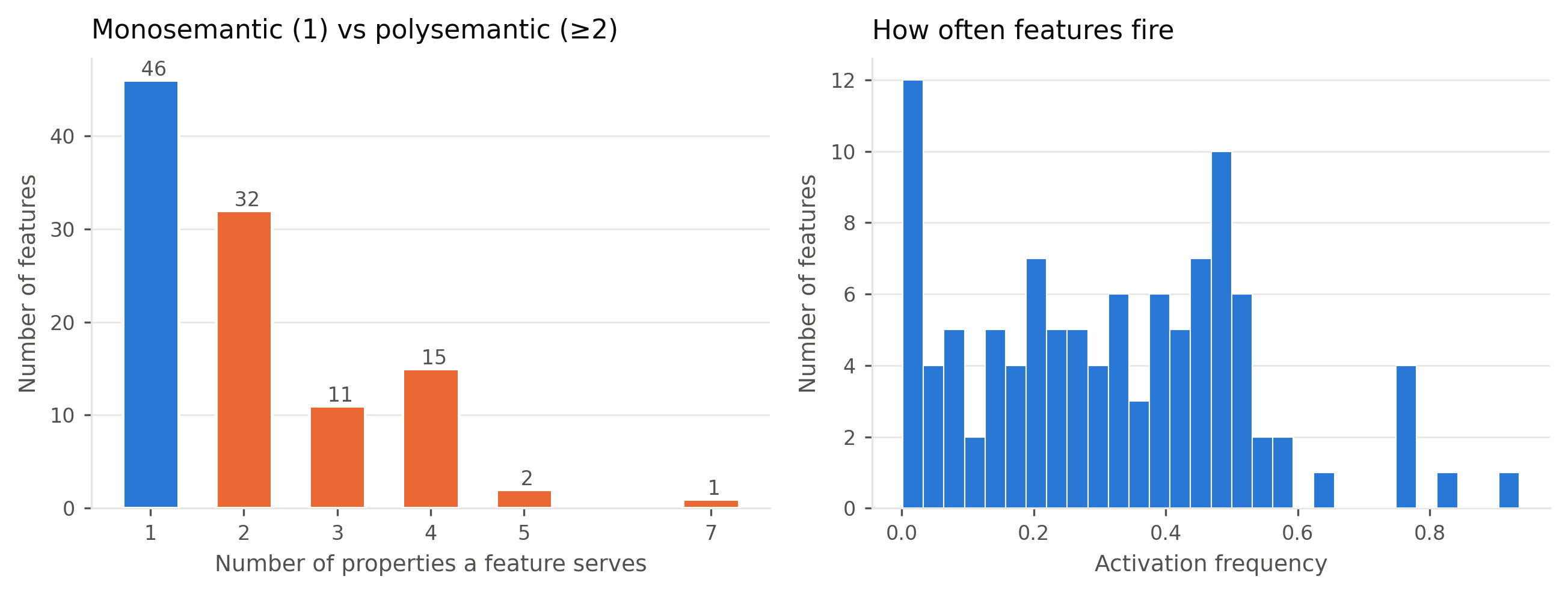}
    \caption{Left: how many targets each top-weighted overcomplete feature serves (blue: single target; orange: two or more). Right: the distribution of activation frequency across the 107 top-weighted features (not the full dictionary, whose mean firing rate is $64/1024\approx6\%$).}
    \label{fig:f8}
\end{figure}

Across the eleven probe weight vectors, 107 features appear in at least one target's top-20 list: 46 (43\%) appear for a single target and 61 (57\%) for two or more (Fig.~\ref{fig:f8}). Because size and molecular weight are the same underlying variable (Section~\ref{sec:decodability}), these eleven nominal targets are ten distinct ones, which slightly inflates the multi-target counts; we report the tallies as computed for comparability but do not read them as evidence of genuine polysemanticity. We describe the single-target group as monosemantic only in the sense that they appear in one target's weight list; this is a statement about probe weights, not proof of a single underlying concept. Substructure enrichment gives 3{,}959 tests, of which 280 are significant (gap above 30 points, $q<0.01$). The strongest are clean, single-halogen detectors: in a representative run, separate features fire almost only on iodine-, chlorine-, bromine-, and fluorine-containing molecules (100\% versus 0 to 17\% in silent molecules, $q<10^{-9}$). One feature detects boron and boronic esters (90\% versus 0\%, $q=5\times10^{-11}$), which is notable because boron appears in only 0.3\% of this corpus, so the enrichment appears despite boron's low corpus prevalence. This enrichment result, from a balanced 30-versus-30 comparison, is the basis for the boron claim rather than the boron probe AUROC, which rests on roughly nine test positives ($92$ across the corpus) and is correspondingly fragile. In this run the SAE thus separates halogens by identity rather than merely flagging halogen presence. The dictionary also contains a clean \emph{absence} detector: one feature fires on molecules lacking a benzene ring (benzene present in 0\% of its top activators versus 93\% of silent molecules, $q=7\times10^{-12}$), so the SAE encodes structural absence as well as presence. Because individual feature indices are not portable across runs (Section~\ref{sec:stability}), we report these patterns rather than relying on the specific indices, which appear in the figures only for concreteness. For drug-likeness, a small number of features carry the largest negative probe weight, matching the structure of the undercomplete analysis; the magnitudes are near $-0.1$ (Ridge on standardized features), so we report ranks and enrichments rather than raw weight magnitudes.

\subsection{Feature-ablation specificity in downstream decoding}
\begin{figure}[htbp]
    \centering
    \includegraphics[width=\linewidth,keepaspectratio]{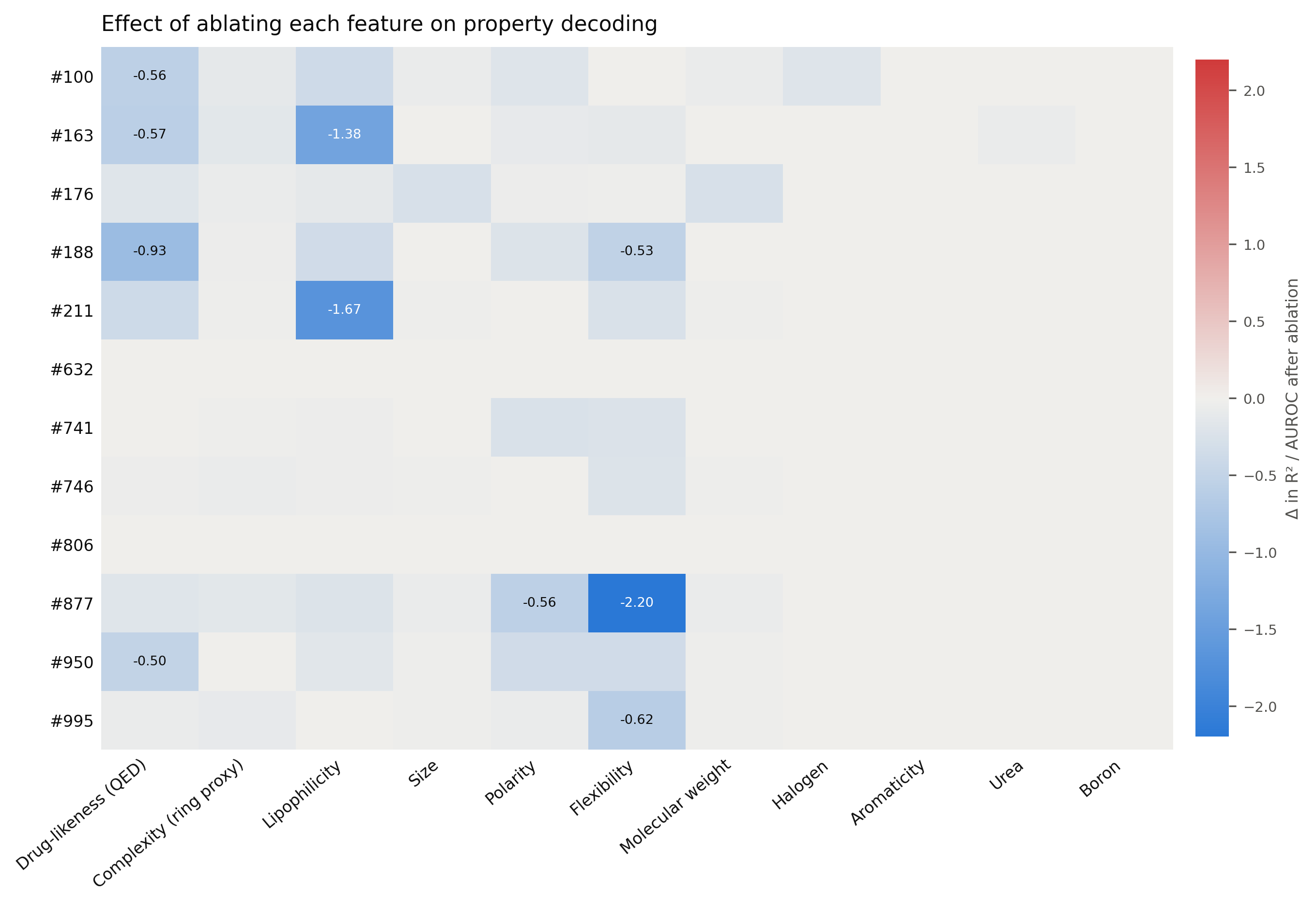}
    \caption{Effect of ablating each feature on property decoding, for a representative run (seed 42). Rows are features, columns are targets, and cell values are the change in $R^2$ or AUROC after the feature is zeroed. Effects concentrate in one column per feature, the signature of specificity. The indices shown are not portable across runs.}
    \label{fig:f7}
\end{figure}

We zero one feature, decode, and re-apply the frozen probes (Fig.~\ref{fig:f7}). The effect is specific: for the clearest features, the targeted property degrades far more than the mean off-target property, with specificity ratios of roughly 8 to 23 times for the substantial interventions and one small intervention reaching $44\times$. We report this ratio rather than the raw change alone, because a large raw change partly reflects general reconstruction damage from zeroing a high-magnitude feature; if the effect were non-specific, off-target properties would degrade in proportion, and these ratios show they do not. The ratio and the effect size must be read together: the highest ratio ($44\times$) comes from a small intervention (targeted $\Delta R^2 \approx -0.014$ against an off-target mean near $0.0003$), whereas the most substantial interventions sit in the $8$ to $23\times$ band with large targeted effects (for example $\Delta R^2 \approx -1.67$ at $23\times$ and $-2.20$ at $18\times$). The band where both magnitude and specificity are large is the transferable claim. This is an interventional effect at the level of probe decoding, not at the level of generation, which was not re-run; it does not by itself establish a causal effect on policy behavior. As with enrichment, the specific indices are from one seed; the transferable claim is the distribution of specificity ratios.

\subsection{Feature stability across runs}
\label{sec:stability}
\begin{figure}[htbp]
    \centering
    \includegraphics[width=0.8\linewidth,keepaspectratio]{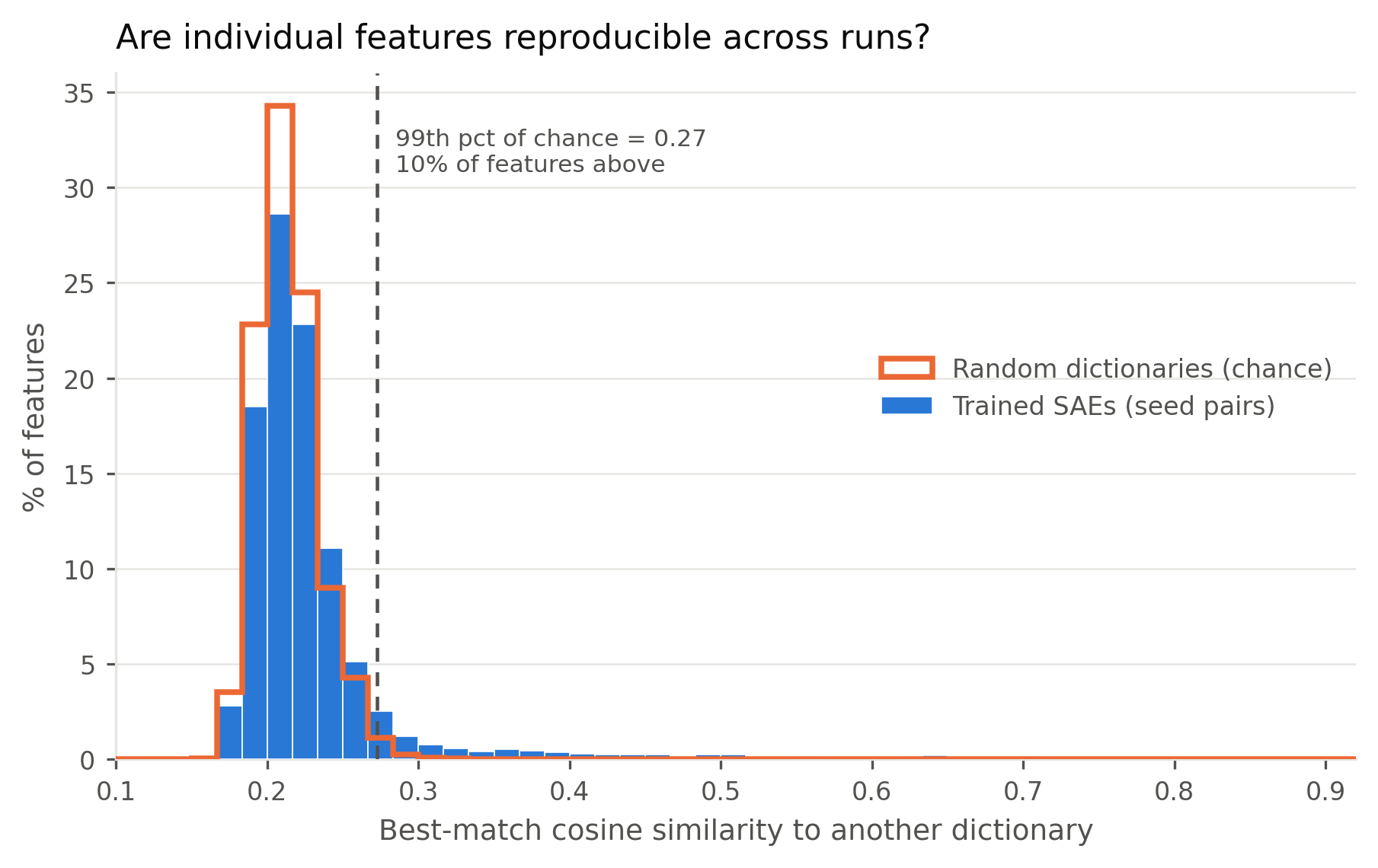}
    \caption{Are individual features reproducible across runs? Best-match cosine similarity between decoder dictionaries from independently seeded runs (blue) against a chance baseline from random dictionaries (orange). Most features sit at chance; a small tail exceeds thresholds chance never reaches.}
    \label{fig:f6}
\end{figure}

Individual features are mostly not reproducible across runs (Fig.~\ref{fig:f6}). Matching decoder columns across independently seeded runs gives a mean best-match cosine of 0.237, against a chance baseline of 0.214 from random dictionaries, only $1.11\times$ chance. The bulk of the dictionary is at chance, but a small core is real: 2.71\% of features exceed cosine 0.5 and 0.97\% exceed 0.7, thresholds no random pair attains. Two consequences follow. First, feature indices are not portable; a named index from one run does not exist in another, and any previously reported index cannot be recovered by retraining. Second, feature-level claims should be restricted to this reproducible core, and the stability check should be reported.

\section{Discussion}

Our results show that interpretability methods such as gradients, saliency, sparse autoencoders, and counterfactual analysis can be adapted to structured generative models. The saliency and counterfactual analysis yields atom-level attributions and candidate reward-sensitive motifs, although the counterfactual effect sizes are small and a controlled saliency-versus-random comparison remains future work; the sparse-autoencoder analysis, with its controls and ablations, provides the more robust evidence. The atom-level attributions and motif detectors correspond to recognizable chemistry (polarity, lipophilicity, aromaticity, halogenation), but, as the controls below show, this correspondence should not be read as evidence that the trained policy learned that chemistry.

The overcomplete extension adds a control-validated, feature-level account. Several claims are supported. A $4\times$ overcomplete BatchTopK autoencoder reconstructs SynFlowNet embeddings at low NMSE with a batch-mean of 64 active features, and the tested $1024$-dimensional BatchTopK model reconstructs substantially better than the tested $128$-dimensional $\ell_1$ baseline at similar average activity. A median of 98\% of probe-detectable information survives the sparse bottleneck (worst case, flexibility, 83\%), and classification is essentially unaffected. The dictionary contains chemically enriched substructure detectors in individual runs, including per-halogen and boron features, of which a small subset of directions is stable across seeds. Single-feature ablation produces property-specific effects, interventional at the level of probe decoding. Drug-likeness is more linearly accessible in the sparse latent than in the dense embedding.

The single most important result, however, is a negative control. A randomly initialized network of identical architecture decodes every probed property essentially as well as the trained policy: the largest gain from training is $+0.036$ (flexibility), and on drug-likeness training buys only $+0.005$, inside the seed spread, while on molecular size the untrained network is marginally better. Linear probing of physicochemical properties is therefore insensitive to what training accomplished; it characterizes the graph architecture and atom featurization, not learned representation. We ran this control under the random split; the same conclusion should hold under the scaffold-disjoint split, but that is untested. It means the probe-decodability results for this model must be read as a statement about graph features rather than acquired chemistry. Training plainly did something (the trained policy generates high-reward synthesizable molecules and the untrained one cannot); the point is that this probe family cannot see it.

We therefore state plainly what is not supported. We do not claim near-perfect probe scores, regardless of split, show learned chemical understanding, because an untrained network matches them. We do not claim the embedding encodes chemistry beyond cheap descriptors, because the descriptor comparison is split-dependent and, more decisively, an untrained network reproduces the embedding's side of it. We do not treat specific feature indices as meaningful across runs, because only a small core reproduces above chance. We do not claim the model has internalized structure-activity reasoning, because nothing here tests it and this probe family cannot. We make no steering or generation-time controllability claim, because generation was never re-run. The claims that survive are narrow and concrete: the overcomplete dictionary is a good, exact-sparsity decomposition; it contains chemically enriched substructure detectors in individual runs, of which a small subset of directions is stable across seeds; and those features have property-specific interventional effects on probe decoding.

A preliminary internal analysis (not part of the published record) reported several values that this controlled, seed-averaged re-run revises: a much lower overcomplete dead-feature fraction (0.6\% versus a previously noted $\sim$54\%, once resampling was applied correctly), a lower raw drug-likeness decodability ($R^2$ 0.59 versus 0.76, a probe result rather than an SAE quantity), and a single- versus multi-target feature split that inverts (43\% single-target here versus 55\% previously). We report the controlled values throughout. One likely contributor to the drug-likeness difference is the RDKit version: QED is version-sensitive, and we pin rdkit 2023.09.5 (Appendix~\ref{sec:overcomplete_appendix}).

Several limitations remain. First, our study focuses primarily on QED, and extending the analysis to multi objective settings such as synthetic accessibility or binding affinity will be necessary to assess broader generalization. Second, our sparse-factorization results rely on sparse autoencoders and probes, which impose relatively simple structure on the latent space. Third, the scaffold split was applied to the probes but not to SAE training: the SAE is unsupervised and sees all 32{,}054 embeddings, which is standard, but a fully scaffold-held-out SAE is untested. Fourth, the untrained-network control uses a single random initialization (probe seeds vary, but the network seed does not), so the small trained-versus-untrained gaps carry probe-split variance rather than initialization variance; the conclusion is robust (a $+0.005$ gap will not become significant under reinitialization), but strictly it is $n=1$ in the controlled factor, and the untrained control was run under the random split only. Fifth, the clean per-halogen and boron detectors were derived from the \emph{trained} embeddings; whether an SAE trained on the untrained network's embeddings would yield the same detectors is the obvious next control and is untested. Sixth, the counterfactual saliency effect sizes are small, the saliency-versus-random test is future work, and the two illustrative $\Delta\mathrm{QED}$ values could not be independently re-verified because the checkpoint is gone.

In line with recent discussions \cite{cretu2025synflownet}, an intriguing direction for future work is to condition GFlowNets directly on physicochemical properties rather than applying post hoc factorization. Such conditioning may yield latent representations that are more naturally aligned with domain relevant chemical axes and may improve controllability in generative design.

\section{Conclusion}
We introduced an interpretability framework for SynFlowNet spanning atom-level saliency, an undercomplete factor analysis, and an overcomplete BatchTopK dictionary, and we validated the probe and dictionary analyses against descriptor, random-label, scaffold-disjoint, untrained-network, ablation, and cross-seed controls (the saliency and counterfactual analysis is illustrative, with its controlled test left as future work). The dictionary contains chemically enriched substructure detectors in individual runs, including separate per-halogen features and a boron detector that appears despite boron's low corpus prevalence, and a small subset of dictionary directions is stable across seeds; single-feature ablation has property-specific effects on probe decoding. Equally important is what the controls rule out. The decisive finding is that an untrained network of identical architecture decodes every probed property essentially as well as the trained policy, so linear probing of physicochemical properties reflects the graph architecture and atom featurization rather than learned representation, and near-perfect probe scores are not evidence of acquired chemistry regardless of split. Individual feature indices also do not survive retraining. We therefore restrict feature-level claims to the reproducible core, present decodability as a property of the architecture rather than of training, and make no steering or controllability claim. Future work should test whether the clean substructure detectors persist when the SAE is trained on untrained-network embeddings, extend the framework to multi-objective GFlowNets, and explore conditioning generative policies directly on physicochemical properties or on interpretable SAE-derived factors.

\section*{Acknowledgements}

We thank our colleagues at Montai for their feedback and support throughout this work. We also acknowledge the use of OpenAI’s ChatGPT to assist with editing and refining the manuscript text.

\bibliographystyle{unsrtnat}
\bibliography{reference}

@inproceedings{cretu2025synflownet,
  title     = {{SynFlowNet}: Design of Diverse and Novel Molecules with Synthesis Constraints},
  author    = {Cretu, Miruna and Harris, Charles and Igashov, Ilia and Schneuing, Arne and Segler, Marwin and Correia, Bruno and Roy, Julien and Bengio, Emmanuel and Li{\`o}, Pietro},
  booktitle = {Proceedings of the International Conference on Learning Representations (ICLR)},
  year      = {2025},
  note      = {arXiv:2405.01155},
}

@article{bickerton2012qed,
  title     = {Quantifying the chemical beauty of drugs},
  author    = {Bickerton, G. Richard and Paolini, Gaia V. and Besnard, J{\'e}r{\'e}my and Muresan, Sorel and Hopkins, Andrew L.},
  journal   = {Nature Chemistry},
  volume    = {4},
  number    = {2},
  pages     = {90--98},
  year      = {2012},
  doi       = {10.1038/nchem.1243},
}

@misc{karimi2020algorithmicrecoursecounterfactualexplanations,
  title         = {Algorithmic Recourse: From Counterfactual Explanations to Interventions},
  author        = {Karimi, Amir-Hossein and Sch{\"o}lkopf, Bernhard and Valera, Isabel},
  year          = {2020},
  eprint        = {2002.06278},
  archivePrefix = {arXiv},
  primaryClass  = {cs.LG},
  url           = {https://arxiv.org/abs/2002.06278},
}

@inproceedings{shu2025surveysparseautoencodersinterpreting,
  title     = {A Survey on Sparse Autoencoders: Interpreting the Internal Mechanisms of Large Language Models},
  author    = {Shu, Dong and Wu, Xuansheng and Zhao, Haiyan and Rai, Daking and Yao, Ziyu and Liu, Ninghao and Du, Mengnan},
  booktitle = {Findings of the Association for Computational Linguistics: EMNLP 2025},
  year      = {2025},
}

@inproceedings{bengio2021gflownet,
  title     = {Flow Network Based Generative Models for Non-Iterative Diverse Candidate Generation},
  author    = {Bengio, Emmanuel and Jain, Moksh and Korablyov, Maksym and Precup, Doina and Bengio, Yoshua},
  booktitle = {Advances in Neural Information Processing Systems (NeurIPS)},
  year      = {2021},
}

@inproceedings{madan2023gflownet,
  title     = {Learning {GFlowNets} from Partial Episodes for Improved Convergence and Stability},
  author    = {Madan, Kanika and Rector-Brooks, Jarrid and Korablyov, Maksym and Bengio, Emmanuel and Jain, Moksh and Nica, Andrei Cristian and Bosc, Tom and Bengio, Yoshua and Malkin, Nikolay},
  booktitle = {Proceedings of the 40th International Conference on Machine Learning (ICML)},
  series    = {PMLR},
  volume    = {202},
  pages     = {23467--23483},
  year      = {2023},
  note      = {arXiv:2209.12782},
}

@inproceedings{sundararajan2017integrated,
  title     = {Axiomatic Attribution for Deep Networks},
  author    = {Sundararajan, Mukund and Taly, Ankur and Yan, Qiqi},
  booktitle = {International Conference on Machine Learning (ICML)},
  year      = {2017},
}

@inproceedings{bau2017network,
  title     = {Network Dissection: Quantifying Interpretability of Deep Visual Representations},
  author    = {Bau, David and Zhou, Bolei and Khosla, Aditya and Oliva, Aude and Torralba, Antonio},
  booktitle = {IEEE Conference on Computer Vision and Pattern Recognition (CVPR)},
  year      = {2017},
}

@inproceedings{ying2019gnnexplainer,
  title     = {{GNNExplainer}: Generating Explanations for Graph Neural Networks},
  author    = {Ying, Rex and Bourgeois, Dylan and You, Jiaxuan and Zitnik, Marinka and Leskovec, Jure},
  booktitle = {Advances in Neural Information Processing Systems (NeurIPS)},
  year      = {2019},
}

@article{jimenezluna2020explainable,
  title     = {Drug discovery with explainable artificial intelligence},
  author    = {Jim{\'e}nez-Luna, Jos{\'e} and Grisoni, Francesca and Schneider, Gisbert},
  journal   = {Nature Machine Intelligence},
  volume    = {2},
  number    = {10},
  pages     = {573--584},
  year      = {2020},
}

@article{wellawatte2022cf,
  title     = {Model agnostic generation of counterfactual explanations for molecules},
  author    = {Wellawatte, Geemi P. and Seshadri, Aditi and White, Andrew D.},
  journal   = {Chemical Science},
  volume    = {13},
  number    = {13},
  pages     = {3697--3705},
  year      = {2022},
  doi       = {10.1039/d1sc05259d},
}

@inproceedings{lucic2022cfgexplainer,
  title     = {{CF-GNNExplainer}: Counterfactual Explanations for Graph Neural Networks},
  author    = {Lucic, Ana and ter Hoeve, Maartje and Tolomei, Gabriele and de Rijke, Maarten and Silvestri, Fabrizio},
  booktitle = {Proceedings of the 25th International Conference on Artificial Intelligence and Statistics (AISTATS)},
  year      = {2022},
}

@article{bricken2023monosemanticity,
  title        = {Towards Monosemanticity: Decomposing Language Models with Dictionary Learning},
  author       = {Bricken, Trenton and Templeton, Adly and Batson, Joshua and Chen, Brian and Jermyn, Adam and Conerly, Tom and Turner, Nicholas and Anil, Cem and Denison, Carson and Askell, Amanda and Lasenby, Robert and Wu, Yifan and Kravec, Shauna and Schiefer, Nicholas and Maxwell, Tim and Joseph, Nicholas and Hatfield-Dodds, Zac and Tamkin, Alex and Nguyen, Karina and McLean, Brayden and Burke, Josiah E. and Hume, Tristan and Carter, Shan and Henighan, Tom and Olah, Christopher},
  journal      = {Transformer Circuits Thread},
  year         = {2023},
  url          = {https://transformer-circuits.pub/2023/monosemantic-features/index.html},
}

@article{cunningham2023sparse,
  title         = {Sparse Autoencoders Find Highly Interpretable Features in Language Models},
  author        = {Cunningham, Hoagy and Ewart, Aidan and Riggs, Logan and Huben, Robert and Sharkey, Lee},
  journal       = {arXiv preprint arXiv:2309.08600},
  year          = {2023},
}

@article{elhage2022superposition,
  title        = {Toy Models of Superposition},
  author       = {Elhage, Nelson and Hume, Tristan and Olsson, Catherine and Schiefer, Nicholas and Henighan, Tom and Kravec, Shauna and Hatfield-Dodds, Zac and Lasenby, Robert and Drain, Dawn and Chen, Carol and Grosse, Roger and McCandlish, Sam and Kaplan, Jared and Amodei, Dario and Wattenberg, Martin and Olah, Christopher},
  journal      = {Transformer Circuits Thread},
  year         = {2022},
}

@article{bussmann2024batchtopk,
  title         = {{BatchTopK} Sparse Autoencoders},
  author        = {Bussmann, Bart and Leask, Patrick and Nanda, Neel},
  journal       = {arXiv preprint arXiv:2412.06410},
  year          = {2024},
}

@article{gao2024scaling,
  title         = {Scaling and Evaluating Sparse Autoencoders},
  author        = {Gao, Leo and la Tour, Tom Dupr{\'e} and Tillman, Henk and Goh, Gabriel and Troll, Rajan and Radford, Alec and Sutskever, Ilya and Leike, Jan and Wu, Jeffrey},
  journal       = {arXiv preprint arXiv:2406.04093},
  year          = {2024},
}

@inproceedings{hewitt2019control,
  title     = {Designing and Interpreting Probes with Control Tasks},
  author    = {Hewitt, John and Liang, Percy},
  booktitle = {Proceedings of the 2019 Conference on Empirical Methods in Natural Language Processing (EMNLP)},
  year      = {2019},
  note      = {arXiv:1909.03368},
}

@article{olshausen1997sparse,
  title     = {Sparse Coding with an Overcomplete Basis Set: A Strategy Employed by {V1}?},
  author    = {Olshausen, Bruno A. and Field, David J.},
  journal   = {Vision Research},
  volume    = {37},
  number    = {23},
  pages     = {3311--3325},
  year      = {1997},
}

@article{zou2006sparse,
  title     = {Sparse Principal Component Analysis},
  author    = {Zou, Hui and Hastie, Trevor and Tibshirani, Robert},
  journal   = {Journal of Computational and Graphical Statistics},
  volume    = {15},
  number    = {2},
  pages     = {265--286},
  year      = {2006},
}

@article{bemis1996scaffold,
  title     = {The Properties of Known Drugs. 1. Molecular Frameworks},
  author    = {Bemis, Guy W. and Murcko, Mark A.},
  journal   = {Journal of Medicinal Chemistry},
  volume    = {39},
  number    = {15},
  pages     = {2887--2893},
  year      = {1996},
  doi       = {10.1021/jm9602928},
}

\appendix

\section{Code and Data Availability}
\label{sec:availability}
The full implementation, including the overcomplete SAE, linear probes, controls, ablation, and stability scripts, is available at: \\
\url{https://github.com/amirtha-montai/synflownet_public/tree/main/src/interpretability}\\
The environment is pinned (rdkit 2023.09.5, torch 2.1.2, scikit-learn 1.2.2; a 224-package byte-exact freeze in \texttt{env/}), all figures rebuild from CSV outputs without a GPU, and every quoted number is traced to its producing script and source table in \texttt{NUMBERS\_MANIFEST.csv}. Because the analyzed checkpoint no longer exists, the 32{,}054 embeddings are the reproducible artifact and are released in the repository above, together with the untrained-network embeddings and its architecture record.

\section{Probe Target Definitions}
\label{sec:targets}
Table~\ref{tab:targets} defines the eleven linear-probe targets. Classification prevalences on this 32{,}054-molecule set are: aromaticity 90.6\% (29{,}045), halogen 49.9\% (16{,}005), urea 0.95\% (306), boron 0.29\% (92). Urea is 0.95\%, not the 1.7\% cited in an earlier draft; its AUROC of 0.92 rests on 306 positives and carries the widest seed spread of any target ($\pm 0.024$).

\begin{table}[H]
\centering
\small
\caption{The eleven linear-probe targets. ``Dropped from baseline'' lists count columns removed per target because the target is defined by them; the leakage-free descriptor baseline omits these. The baseline never contains MolWt, LogP, or TPSA, so molecular weight, size, polarity, lipophilicity, drug-likeness, and urea have nothing dropped.}
\label{tab:targets}
\begin{tabular}{llll}
\toprule
Target & Type & Definition & Dropped from baseline \\
\midrule
boron           & classification & contains boron (SMARTS \texttt{[\#5]})            & \texttt{n\_B} \\
halogen         & classification & contains F, Cl, Br, or I                          & \texttt{n\_halogen} \\
aromaticity     & classification & contains an aromatic ring                         & ring counts \\
urea            & classification & SMARTS \texttt{[NX3][CX3](=[OX1])[NX3]}           & --- \\
drug-likeness   & regression     & QED                                               & --- \\
molecular weight& regression     & MolWt                                              & --- \\
size            & regression     & clamp(MolWt/500, 0, 2)                             & --- (redundant with MW) \\
polarity        & regression     & clamp(TPSA/200, 0, 2)                              & --- \\
lipophilicity   & regression     & clamp(MolLogP, $-10$, 10)                          & --- \\
flexibility     & regression     & clamp(RotatableBonds/10, 0, 2)                    & \texttt{NumRotatableBonds} \\
complexity      & regression     & ring-count proxy (Appendix~\ref{sec:sae_analysis})& ring counts \\
\bottomrule
\end{tabular}
\end{table}

\section{Sparse Autoencoder Analysis}
\label{sec:sae_analysis}

To better understand how QED (drug-likeness) is represented internally, we trained the undercomplete MLP-SAE on SynFlowNet embeddings. 
The MLP-SAE reveals interpretable chemical axes such as \textit{size}, \textit{polarity}, and \textit{lipophilicity}, suggesting that, in these embeddings, drug-likeness is associated with simpler physicochemical components rather than a single latent dimension.

\subsection{Dataset Summary}
\begin{itemize}
    \item Total molecules analyzed: 32{,}054
    \item Embedding dimension: 256
    \item Number of latent factors discovered: 128
    \item Number of reward signals: 6 (the overcomplete analysis extends this to 11 probe targets)
    \item Train/test split: 28{,}848 / 3{,}206 (90/10, \texttt{random\_state}$=42$)
\end{itemize}

\subsection{Reward Signal Definitions}
The six reward signals are computed per molecule from RDKit descriptors and clamped:
\begin{itemize}
    \item \textbf{drug\_likeness} $= \mathrm{clamp}(\mathrm{QED},0,1)$.
    \item \textbf{complexity} $= 1 - \min\!\big(1.0,\ 0.1\cdot n_{\text{aliphatic rings}} + 0.05\cdot n_{\text{aromatic rings}} + 0.5\big)$, a clipped two-ring-count proxy. It is \textbf{not} the Ertl-Schuffenhauer synthetic-accessibility score and not Bertz complexity; the corresponding $R^2$ measures recovery of this proxy. Note the value \emph{decreases} with ring count.
    \item \textbf{lipophilicity} $= \mathrm{clamp}(\mathrm{MolLogP},-10,10)$.
    \item \textbf{size} $= \mathrm{clamp}(\mathrm{MolWt}/500,0,2)$ (a rescaling of molecular weight).
    \item \textbf{polarity} $= \mathrm{clamp}(\mathrm{TPSA}/200,0,2)$.
    \item \textbf{flexibility} $= \mathrm{clamp}(n_{\text{rotatable bonds}}/10,0,2)$.
\end{itemize}

\subsection{Training Summary}
\begin{figure}[htbp]
    \centering
    \includegraphics[width=\linewidth,keepaspectratio]{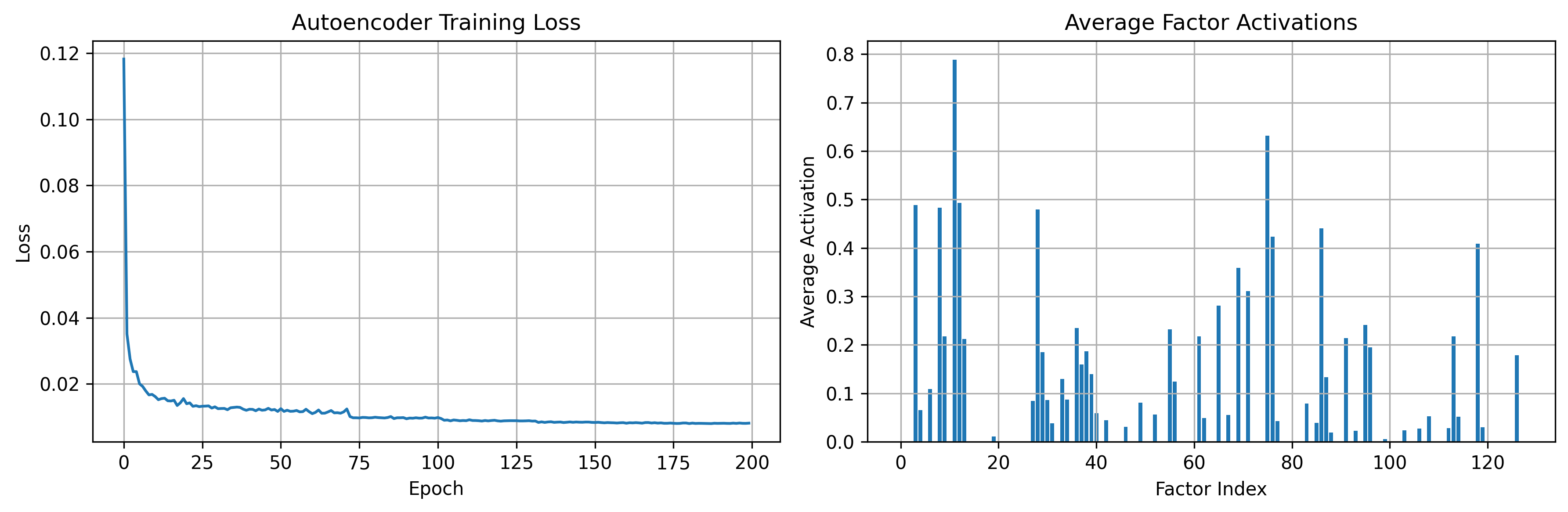}
    \caption{Left: MLP-SAE training loss rapidly decreases and plateaus after $\sim$50 epochs, indicating stable convergence of the reconstruction and sparsity objectives.
Right: Average latent factor activations across 128 neurons show sparse, selective patterns—most factors remain near-zero while a subset exhibits strong activation, consistent with a sparse, selective factorization.}
    \label{fig:sae_training}
\end{figure}

Figure~\ref{fig:sae_training} shows the MLP-SAE training dynamics: the reconstruction-plus-sparsity loss and the distribution of average factor activations across the 128 latent units.

\subsection{Sparsity Analysis}
The MLP-SAE learns a sparse and selective latent representation in which most factors activate for only a small subset of molecules. This pattern indicates that each factor captures a specific and localized chemical feature rather than a broad or entangled signal. A few factors show higher activation frequency, reflecting more general physicochemical dimensions such as polarity or size. Overall, the sparsity distribution shows that the autoencoder yields a compact, property-associated factorization of the embedding.

\begin{figure}[htbp]
    \centering
    \includegraphics[width=\linewidth,keepaspectratio]{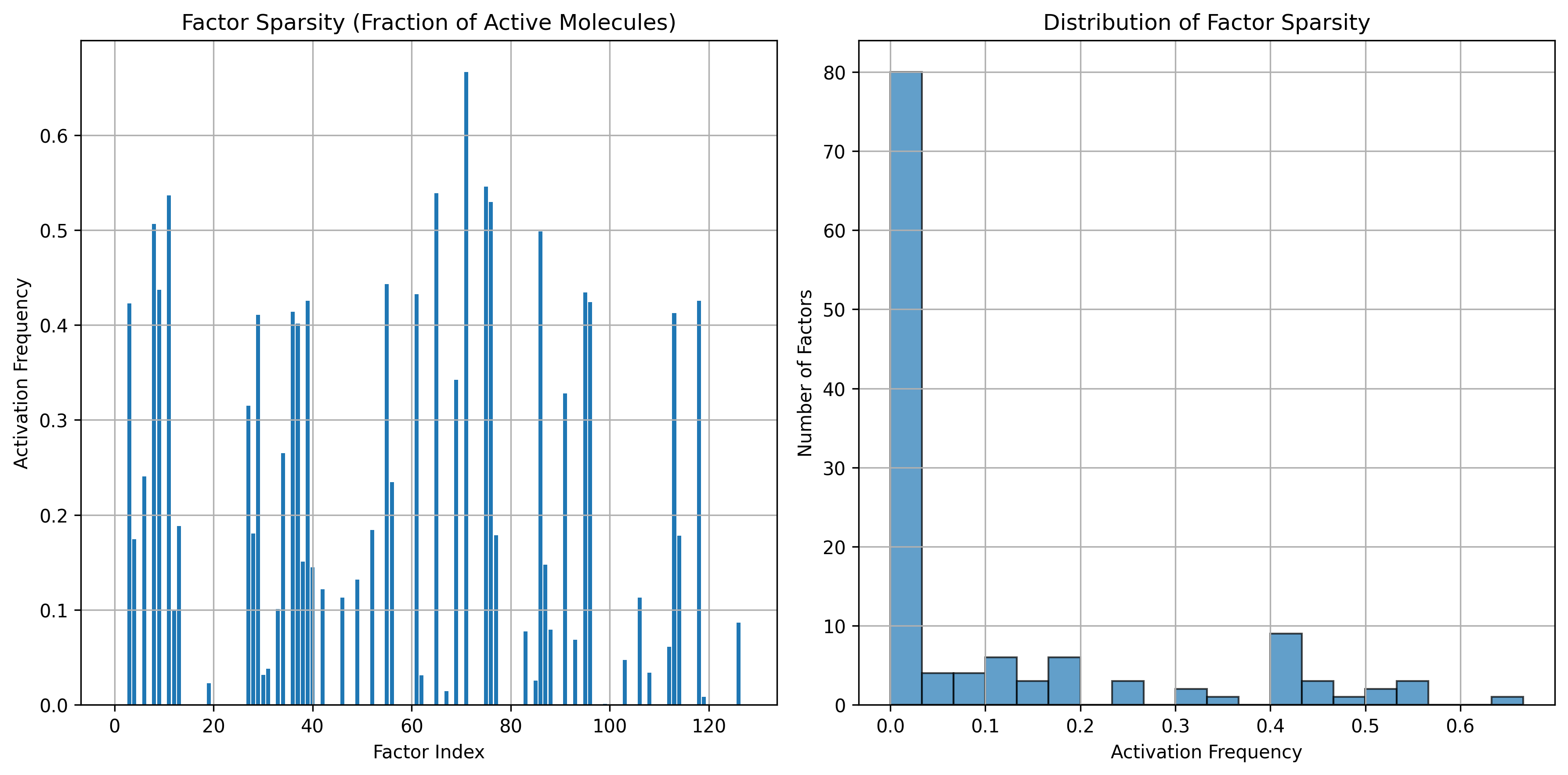}
    \caption{Left: Fraction of active molecules per factor (“activation frequency”) reveals that most latent factors are only triggered by a subset of molecules, while a few are broadly active across the dataset.
Right: Histogram of factor activation frequencies confirms a right-skewed sparsity distribution—over half of the factors activate in fewer than 10\% of molecules—so the autoencoder yields a compact, selective factorization of the embedding.}
    \label{fig:sae_analysis}
\end{figure}

\subsection{Reward Prediction Performance}
Because the relationship between factors and properties might not be purely linear, a three-layer MLP predictor can capture this nonlinearity, giving a more faithful estimate of how well the latent space encodes that property.
A high $R^2$ shows that these properties are accessible from the compressed representation. It does not establish that this structure was acquired through policy training; the untrained-network control (Section~\ref{sec:randomnet}) indicates that much of the decodability is architecture-driven.
\begin{table}[htbp]
\centering
\begin{tabular}{lcc}
\toprule
Reward Signal & Train $R^2$ & Test $R^2$ \\
\midrule
Drug-likeness   & 0.289 & 0.251 \\
Complexity      & 0.756 & 0.750 \\
Lipophilicity   & 0.670 & 0.664 \\
Size            & 0.735 & 0.711 \\
Polarity        & 0.920 & 0.918 \\
Flexibility     & 0.488 & 0.502 \\
\bottomrule
\end{tabular}
\caption{Predictive $R^2$ scores for six chemical reward signals, obtained with a three-layer MLP trained on undercomplete (MLP-SAE) latent factors. Polarity and size are well captured ($R^2 > 0.7$), whereas composite drug-likeness (QED) is harder to predict directly, suggesting that interpretable components underpin QED. ``Complexity'' is the ring-count proxy defined above, not synthetic accessibility.}
\end{table}

\subsection{Latent Factor Sparsity}
Sparsity statistics (fraction of molecules with factor activation $>$ 0.1):  
\begin{itemize}
    \item Mean: 0.105
    \item Std: 0.171
    \item Min: 0.000
    \item Max: 0.666
\end{itemize}

\subsection{Top Factor–Reward Associations}
Table~\ref{tab:topfactor} lists the latent factors most strongly correlated with each reward signal, ranked by absolute Pearson correlation.
\begin{table}[H]
\centering
\begin{tabular}{lcl}
\toprule
Factor & Reward Signal & Correlation (r) \\
\midrule
Factor\_11  & Size        &  0.757 \\
Factor\_75  & Size        & -0.574 \\
Factor\_86  & Polarity    & -0.570 \\
Factor\_118 & Polarity    &  0.540 \\
Factor\_28  & Size        &  0.525 \\
Factor\_12  & Size        & -0.507 \\
Factor\_52  & Polarity    &  0.446 \\
Factor\_49  & Size        & -0.422 \\
Factor\_34  & Polarity    & -0.415 \\
Factor\_96  & Complexity  & -0.412 \\
\bottomrule
\end{tabular}
\caption{Latent factors extracted by the undercomplete MLP-SAE and their strongest correlations with chemical reward signals. Several factors align with interpretable physicochemical properties such as size (Factor\_11) and polarity (Factor\_86, Factor\_118).}
\label{tab:topfactor}
\end{table}

\subsection{Reward-Specific Factor Summary}
\begin{itemize}
    \item Drug-likeness: Factor\_28 (0.379), Factor\_62 (0.325), Factor\_11 (0.310)
    \item Complexity: Factor\_96 (0.412), Factor\_29 (0.380), Factor\_86 (0.365)
    \item Lipophilicity: Factor\_11 (0.412), Factor\_86 (0.387), Factor\_118 (0.355)
    \item Size: Factor\_11 (0.757), Factor\_75 (0.574), Factor\_28 (0.525)
    \item Polarity: Factor\_86 (0.570), Factor\_118 (0.540), Factor\_52 (0.446)
    \item Flexibility: Factor\_55 (0.375), Factor\_11 (0.334), Factor\_36 (0.300)
\end{itemize}

\subsection{Factor–reward correlations}
\begin{figure}[htbp]
    \centering
    \includegraphics[width=1.0\textwidth,height=0.9\textheight,keepaspectratio]{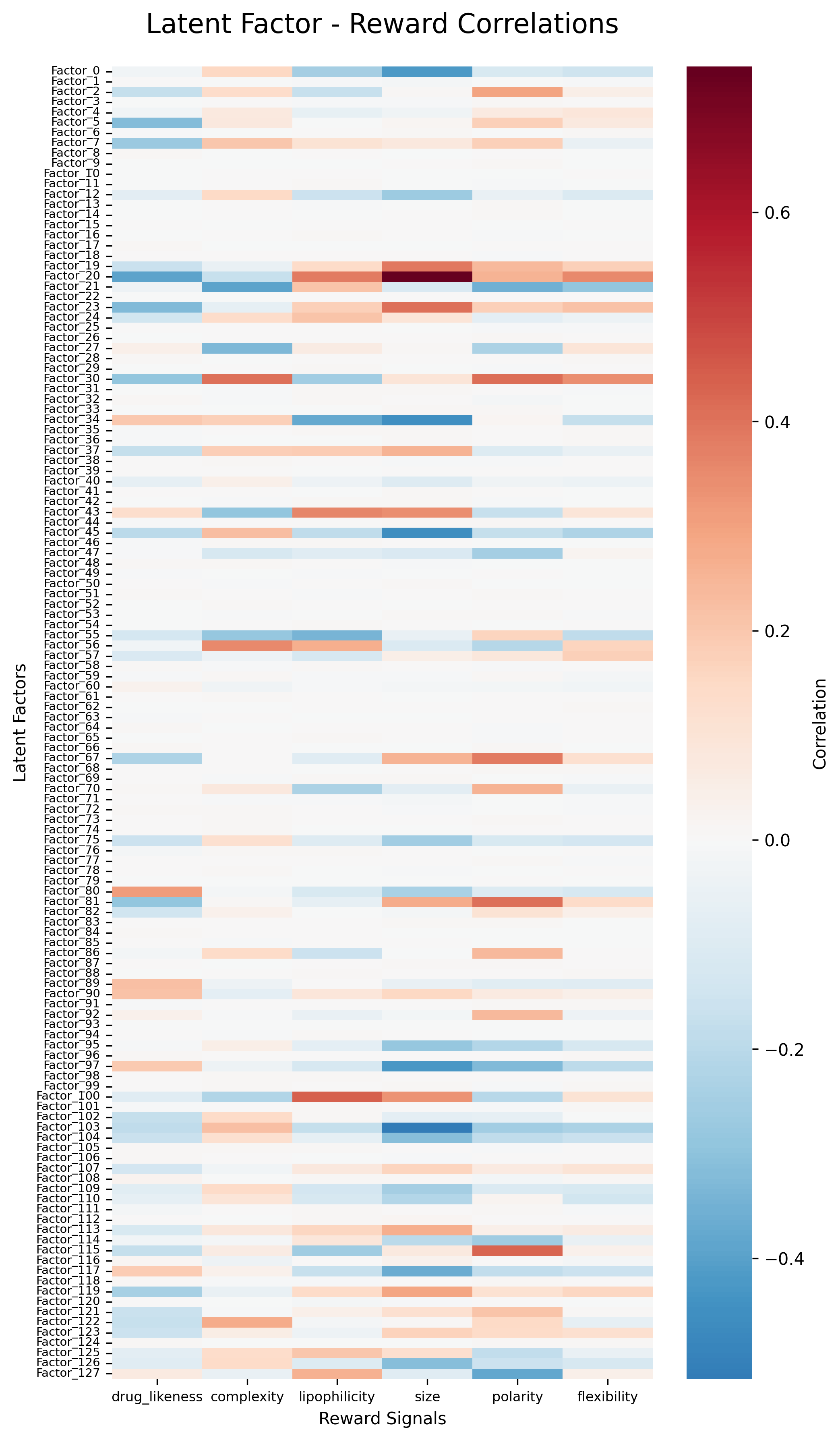}
    \caption{Factor–reward correlation heatmap.}
    \label{fig:corr_factor}
\end{figure}

This analysis aims at comparing each individual latent factor vs.\ one reward. As shown in Figure~\ref{fig:corr_factor}, several latent factors align strongly with polarity and size.

\newpage

\section{Additional Motif Probe Results}
\label{sec:motif_appendix}

\subsection{Training Summary of motifs}
\begin{figure}[htbp]
    \centering
    \includegraphics[width=\linewidth,keepaspectratio]{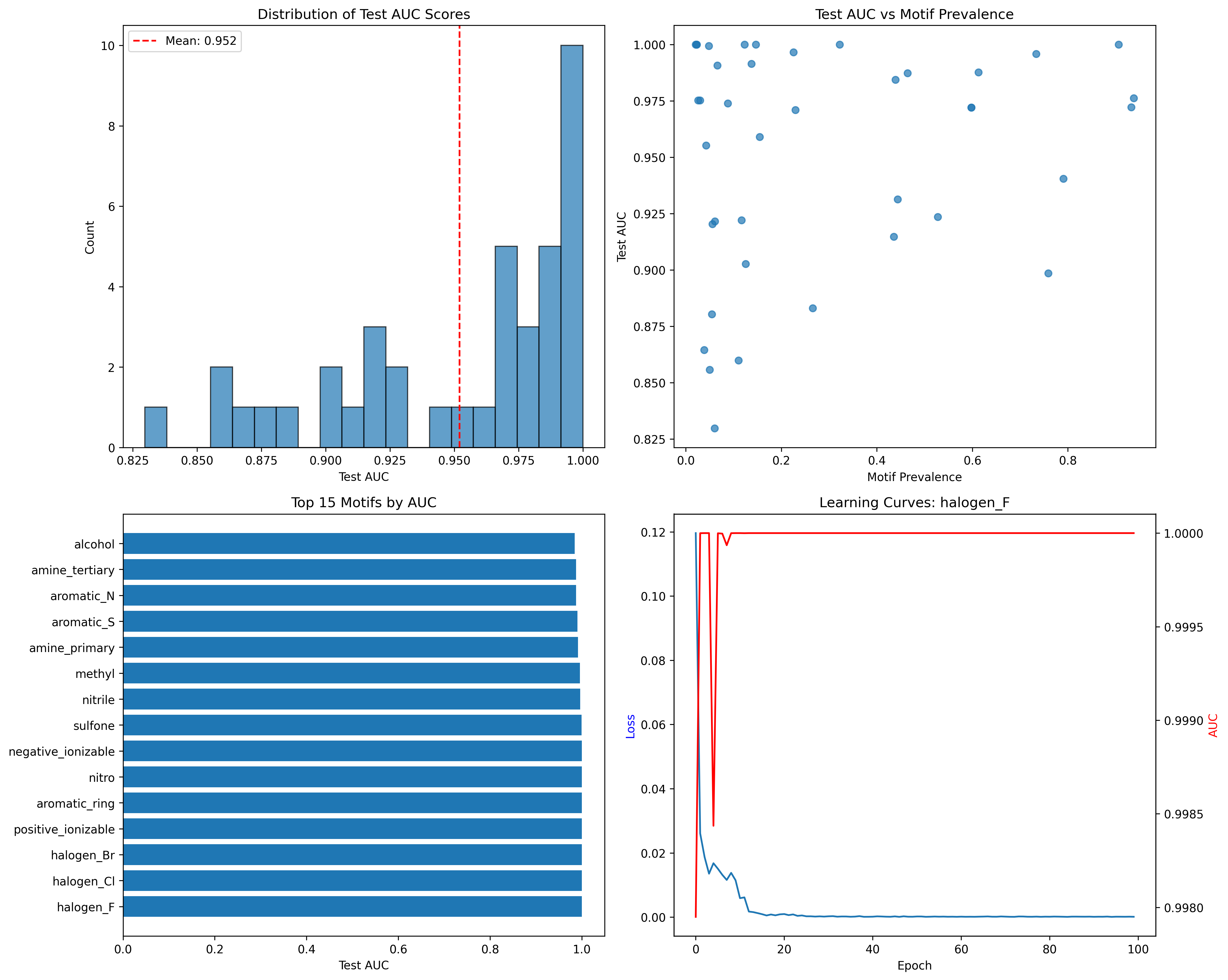}
    \caption{Motif probe results.}
    \label{fig:motif_probe_results}
\end{figure}

Figure~\ref{fig:motif_probe_results} shows that SynFlowNet embeddings encode fine-grained functional group information that maps onto chemically interpretable motifs.

The mean AUC is high ($\approx$ 0.95) with no strong prevalence dependence and stable learning dynamics, so motif information is accessible to shallow classifiers. Whether that accessibility depends on training is untested here: the untrained-network control of Section~\ref{sec:randomnet} was run for the physicochemical probes but not for the motif probes, and Section~\ref{sec:decodability} shows several simple motifs reach AUROC near 1.000 because their atom-type information is directly available in the graph. These scores are therefore subject to the same discount, and control baselines are required before they are attributed to learned representation.

Table~\ref{tab:motif_results} summarizes AUROC and average precision (AP) scores for motif classification using motif probes across all tested motifs.

\begin{center}
\small
\begin{longtable}{@{}lccc@{}}
\caption{Motif probe classification results across diverse functional groups. High \textit{AUROC} ($>$ 0.9) for halogens, aromatic rings, and ionizable groups shows that motif information is accessible to shallow classifiers; for the highest scores, see the trivial-decodability controls in Section~\ref{sec:decodability}.}
\label{tab:motif_results}\\
\toprule
Motif & Prevalence & AUROC & AP \\
\midrule
\endfirsthead

\multicolumn{4}{l}{\small\textit{Table \ref{tab:motif_results} (continued)}}\\
\toprule
Motif & Prevalence & AUROC & AP \\
\midrule
\endhead

\bottomrule
\endfoot

\bottomrule
\endlastfoot

positive\_ionizable & 0.0226 & 1.0000 & 1.0000 \\
halogen\_F          & 0.3222 & 1.0000 & 1.0000 \\
halogen\_Br         & 0.1229 & 1.0000 & 1.0000 \\
aromatic\_ring      & 0.9061 & 1.0000 & 1.0000 \\
halogen\_Cl         & 0.1467 & 1.0000 & 1.0000 \\
nitro               & 0.0208 & 0.99998 & 0.99927 \\
negative\_ionizable & 0.0221 & 0.99998 & 0.99924 \\
sulfone             & 0.0484 & 0.99941 & 0.98622 \\
nitrile             & 0.2253 & 0.99660 & 0.98899 \\
methyl              & 0.7334 & 0.99588 & 0.99833 \\
amine\_primary      & 0.1375 & 0.99146 & 0.95530 \\
aromatic\_S         & 0.0658 & 0.99069 & 0.87042 \\
aromatic\_N         & 0.6129 & 0.98763 & 0.99259 \\
amine\_tertiary     & 0.4640 & 0.98726 & 0.98557 \\
alcohol             & 0.4387 & 0.98442 & 0.98150 \\
methylene           & 0.9379 & 0.97621 & 0.99842 \\
terminal\_alkyne    & 0.0258 & 0.97523 & 0.70965 \\
thiophene           & 0.0296 & 0.97522 & 0.48240 \\
aromatic\_O         & 0.0878 & 0.97386 & 0.83362 \\
hbond\_acceptor     & 0.9326 & 0.97217 & 0.99793 \\
phenyl              & 0.5976 & 0.97210 & 0.97911 \\
benzene             & 0.5976 & 0.97199 & 0.97888 \\
carboxylic\_acid    & 0.2296 & 0.97099 & 0.90775 \\
quaternary\_carbon  & 0.1549 & 0.95900 & 0.83732 \\
triple\_bond        & 0.0422 & 0.95524 & 0.65181 \\
hbond\_donor        & 0.7903 & 0.94051 & 0.98364 \\
ether               & 0.4434 & 0.93137 & 0.91106 \\
amide               & 0.5278 & 0.92347 & 0.92691 \\
ester               & 0.1167 & 0.92204 & 0.60984 \\
double\_bond        & 0.0610 & 0.92153 & 0.56933 \\
vinyl               & 0.0559 & 0.92028 & 0.53539 \\
amine\_secondary    & 0.4357 & 0.91477 & 0.89198 \\
cyclohexane         & 0.1252 & 0.90263 & 0.57151 \\
methine             & 0.7588 & 0.89854 & 0.96505 \\
pyridine            & 0.2656 & 0.88299 & 0.73224 \\
cyclopentane        & 0.0547 & 0.88028 & 0.36046 \\
allyl               & 0.0384 & 0.86442 & 0.33325 \\
cyclopropane        & 0.1104 & 0.85982 & 0.42632 \\
phenol              & 0.0499 & 0.85573 & 0.25594 \\
ketone              & 0.0604 & 0.82974 & 0.36870 \\
\end{longtable}
\end{center}

\subsection{Ground Truth Motif correlation}
\begin{figure}[htbp]
    \centering
    \includegraphics[width=\linewidth,keepaspectratio]{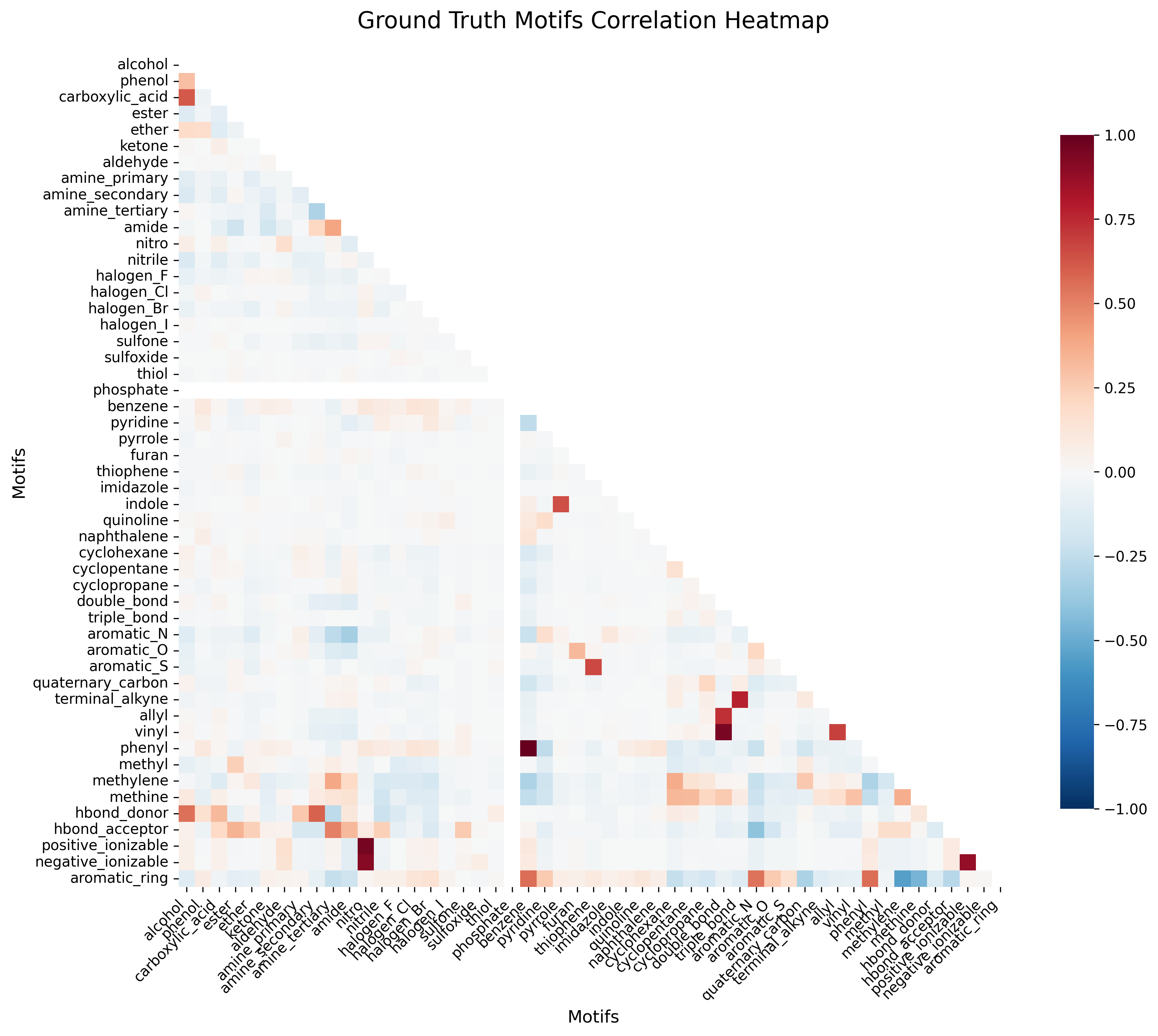}
    \caption{Ground-truth motif–motif correlations.}
    \label{fig:motif_gt_correlation}
\end{figure}

Figure~\ref{fig:motif_gt_correlation} serves as a qualitative baseline for embedding interpretability:
if SynFlowNet’s motif–motif correlation heatmap (from learned probes) matches this ground-truth pattern, it indicates that the model captures real chemical dependencies rather than spurious ones. Table~\ref{tab:motif_comparison} summarizes the qualitative comparison; we report it descriptively and attach no statistic to it.

\begin{table}[H]
\centering
\small
\caption{Qualitative comparison between the ground-truth motif–motif correlations (Figure~\ref{fig:motif_gt_correlation}) and the SynFlowNet-extracted structure. Descriptive only.}
\begin{tabular}{p{4cm}p{4.5cm}p{4.5cm}}
\toprule
\textbf{Aspect} & \textbf{Ground Truth} & \textbf{Extracted from SynFlowNet} \\
\midrule
Aromatic and halogen clusters & Strong, localized & Strong, slightly enhanced \\
Polar motif structure & Moderate, noisy & Clearer, cohesive \\
Negative correlations & Present (polar vs.\ nonpolar) & Preserved \\
Rare motif relationships & Sparse & Smoother, more correlated \\
\bottomrule
\end{tabular}
\label{tab:motif_comparison}
\end{table}

\section{Overcomplete SAE: Environment and Hyperparameter Sweep}
\label{sec:overcomplete_appendix}
The overcomplete analysis uses rdkit 2023.09.5 (pinned, since QED is version-sensitive), torch 2.1.2, scikit-learn 1.2.2, on a single Tesla T4. Autoencoder seeds are 42 to 46 (dictionary initialization) and probe seeds are 0 to 4 (data splits); both are reported independently. Before selecting BatchTopK with $k=64$ and dead-neuron resampling, we swept $\ell_1$ (ReLU) and BatchTopK families at dictionary sizes 1024 and 2048, following the evaluation practice of Gao et al.~\cite{gao2024scaling}. ReLU with $\ell_1$ in $[10^{-3},10^{-2}]$ reconstructed well but failed to reach meaningful sparsity ($L_0>500$), while BatchTopK with small $k$ (8, 16) enforced exact sparsity but produced 85 to 97\% dead features on this dataset size. The chosen configuration balances near-exact reconstruction, meaningful sparsity ($L_0=64$), and negligible feature death under resampling.

\end{document}